%% file: acl_latex.tex
\documentclass[11pt]{article}

\usepackage[final]{acl}
\usepackage{amssymb}
\usepackage{times}
\usepackage{latexsym}
\usepackage{amsmath}
\usepackage{subcaption} 
\usepackage{mdframed}
\usepackage{amsthm}
\usepackage[most]{tcolorbox}
\usepackage{lipsum}  % 用于测试文本，可删除

\tcbset{
  enhanced,
  colback=gray!5,
  colframe=black!60,
  boxrule=0.5pt,
  sharp corners,
  fonttitle=\bfseries,
  coltitle=black,
  left=6pt,
  right=6pt,
  top=6pt,
  bottom=6pt,
  width=\textwidth,
  enlarge left by=0mm,
  enlarge right by=0mm,
}

\newtheorem{finding}{Finding}
\usepackage[T1]{fontenc}
\usepackage[utf8]{inputenc}
\usepackage{arydshln}
\usepackage{microtype}

\usepackage{inconsolata}

\usepackage{multirow}
\usepackage{graphicx}
\usepackage{booktabs}
\usepackage{makecell}   % 提供\multirowcell
\usepackage{enumitem}

\title{From Sub-Ability Diagnosis to Human-Aligned Generation: Bridging the Gap for Text Length Control via \textsc{MarkerGen}}
\author{Peiwen Yuan$^1$\footnotemark[1], Chuyi Tan$^1$\footnotemark[1], Shaoxiong Feng$^2$, Yiwei Li$^1$, Xinglin Wang$^1$\\ 
  {\bf Yueqi Zhang$^1$, Jiayi Shi$^1$, Boyuan Pan$^2$, Yao Hu$^2$, Kan Li$^{1}$\footnotemark[2]}\\
  $^1$School of Computer Science and Technology, Beijing Institute of Technology \\
  $^2$Xiaohongshu Inc \\
\texttt{\{peiwenyuan,tanchuyi,liyiwei,wangxinglin,zhangyq,shijiayi,likan\}@bit.edu.cn} \\
\texttt{\{shaoxiongfeng2023,whd.thu\}@gmail.com} \ \
\texttt{\{panboyuan,xiahou\}@xiaohongshu.com}}

\begin{document}
\maketitle
\renewcommand{\thefootnote}{\fnsymbol{footnote}} 
\footnotetext[1]{Equal contribution.} 
\footnotetext[2]{Corresponding author.} 

\renewcommand{\thefootnote}{\arabic{footnote}}

\input{abs}
\input{intro}
\input{prelimilary}
\input{method}

\input{exp}

\input{conclusion}
\bibliography{acl_latex.bbl}
\appendix

\input{appendix}

\end{document}

%% file: abs.tex
\begin{abstract}
% Significant progress has been made in automatic text evaluation with the introduction of large language models (LLMs) as evaluators. However, current sample-wise evaluation paradigm suffers from the following issues: (1) Sensitive to prompt design; (2) Poor resistance to noise; (3) Inferior ensemble performance with static reference. Inspired by the fact that humans treat both criterion definition and inter sample comparison as references for evaluation, we propose BATCHEVAL, a paradigm that conducts batch-wise evaluation iteratively to alleviate the above problems. We explore variants under this paradigm and confirm the optimal settings are two stage procedure with heterogeneous batch composition strategy and decimal scoring format. Comprehensive experiments across 3 LLMs on 4 text evaluation tasks demonstrate that BATCHEVAL outperforms state-of-the-art methods by 10.5\% on Pearson correlations with only 64\% API cost on average. Further analyses have been conducted to verify the robustness, generalization, and working mechanism of BATCHEVAL.
% 大语言模型(LLMs)的快速发展并没有使其length-controllable text generation（LCTG） 能力达到期望水准，这限制了大量的实际应用需求。
Despite the rapid progress of large language models (LLMs), their length-controllable text generation (LCTG) ability remains below expectations, posing a major limitation for practical applications.
Existing methods mainly focus on end-to-end training to reinforce adherence to length constraints. 
However, the lack of decomposition and targeted enhancement of LCTG sub-abilities restricts further progress.
To bridge this gap, we conduct a bottom-up decomposition of LCTG sub-abilities with human patterns as reference and perform a detailed error analysis.
% 当前的方法端到端地训练模型加强对长度限制的遵循，但由于缺少对LCTG sub-capabilities的decomposition and targeted enhancement，thereby limiting their progress.
% 为了fill this gap，我们首先以人类为参照对LCTG的子能力进行了自底向上的分解，并分别进行了error分析。
% 在此基础上，我们提出了simple yet effective, plug-and-play的\textsc{MakerGen}方法，which：(1) 通过外部工具引用弥补LLM基础能力的不足(2)通过自适应length markers插入来对长度显式建模（3）通过两阶段的生成范式来保证 length requirements are better met without compromising content quality.
On this basis, we propose \textsc{MarkerGen}, a simple-yet-effective plug-and-play approach that:
(1) mitigates LLM fundamental deficiencies via external tool integration;
(2) conducts explicit length modeling with 
dynamically inserted markers;
(3) employs a three-stage generation scheme to better align length constraints while maintaining content quality.
% 我们comprehensive的实验证明\textsc{MakerGen}在多种settings下都极大的提升了LCTG，展现了卓越的有效性和泛化性。
Comprehensive experiments demonstrate that \textsc{MarkerGen} significantly improves LCTG across various settings, exhibiting outstanding effectiveness and generalizability.\footnote{Our code have been released on \url{https://github.com/chuyi369/MarkerGen}.}.
\end{abstract}

%% file: intro.tex
\section{Introduction}
As a fundamental attribute of text generation, ensuring controllability over text length is of great importance \citep{CTGsurvey}. 
% 文本类型（摘要、故事创作等）、用户需求（倾向像是或精简的文本）、外在限制（社交平台的字数限制）的不同，会导向不同的长度需求，which在实际应用中是广泛存在的。
Different text types (e.g., summary, story), user needs (e.g., preference for detailed or concise writing), and external requirements (e.g., social media character limits) shape varied length constraints, which are widely present in real-world scenarios \citep{zhang2023survey}.
% 随着大语言模型发展带来的LLM应用增多，length controllable text generation (LCTG) 变得愈发重要。
With the rapid development of LLMs, their expanding range of applications has made length-controllable text generation (LCTG) even more crucial in current era \citep{foster2024token,gu2024length}.

\begin{figure*}[t]
  % \vspace{-1cm} % 负间距让图片靠近顶部
  \includegraphics[width=\textwidth, height=0.3\textheight]{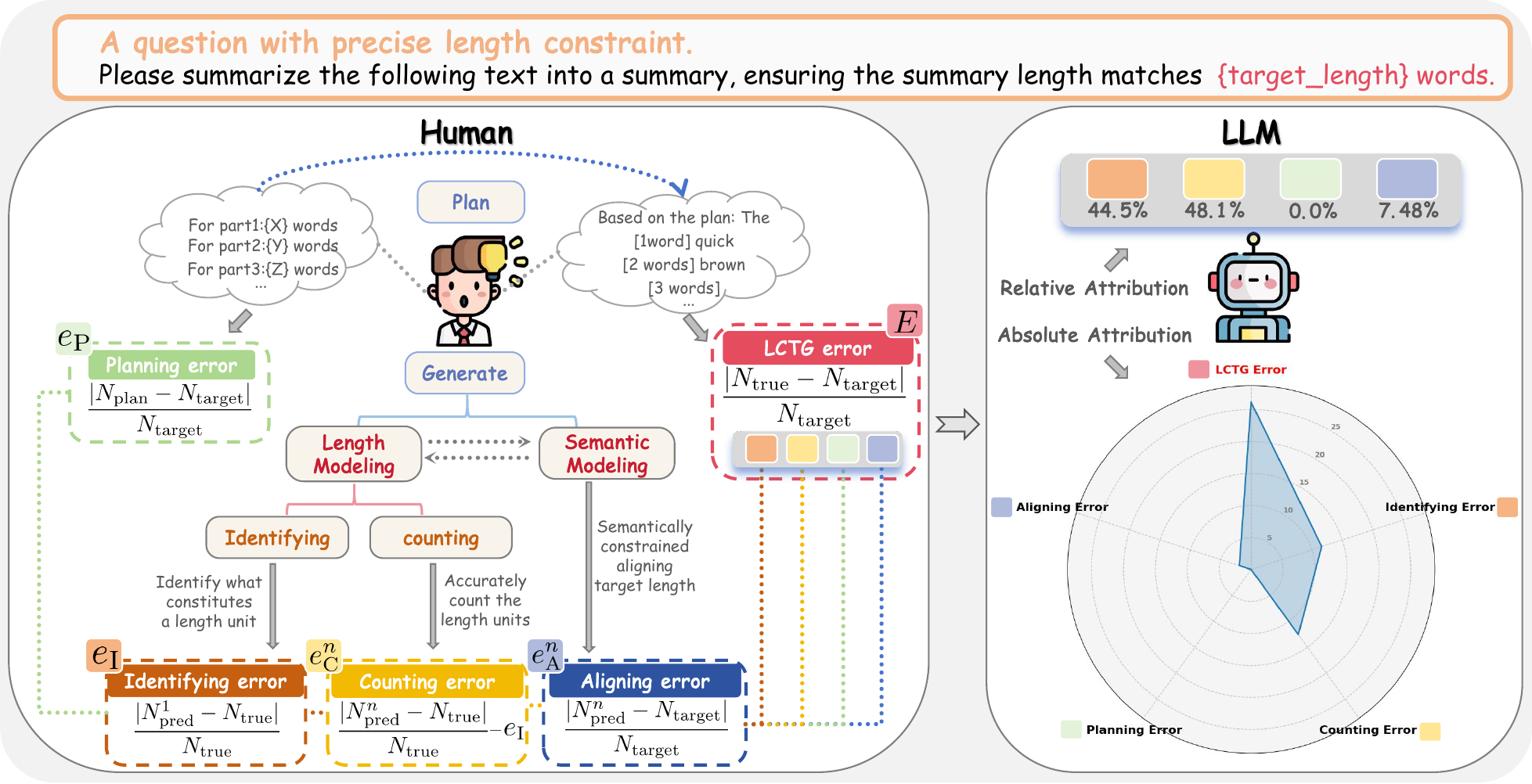}
  \caption{Sub-ability decomposition of LCTG and corresponding error analysis in LLMs.}
  \vspace{-0.5cm} % 负间距让图片靠近顶部
  \label{fig:Figure0}
\end{figure*}
% \begin{figure*}[t]
%   % \vspace{-1.8cm} % 负间距让图片靠近顶部
%   \includegraphics[width=\textwidth, height=0.3\textheight]{fig/Figure0.pdf}
%   \caption{Sub-ability Decomposition of LCTG and Corresponding Error Analysis in LLMs.}
%   \label{fig:Figure0}
% \end{figure*}

% 目前，监督学习方法在广泛的长度尺度上精确建模文本长度方面存在困难，导致控制过于粗略且泛化能力有限。离散的长度监督导致长度遵循不准确，使得细粒度控制变得困难。此外，这些方法依赖于固定的长度分布和指令格式，限制了它们在处理不同提示和领域中的多样化长度约束时的有效性。
% Currently, supervised learning methods struggle to precisely model text length across a broad range of scales, resulting in coarse-grained control and limited generalization. Discrete length supervision leads to imprecise length adherence, making fine-grained control difficult. Moreover, these methods rely on fixed length distributions and instruction formats, limiting their ability to effectively handle diverse length constraints across different prompts and domains.

% 然而，LLM能力的提升并没有使当前的LCTG达到期望的效果。\citet{lift}实验发现当前advanced LLMs(e.g., GPT4-Turbo) violates length constraints in the prompt almost 50% of the time. 一些方法尝试了training-based的方法强化模型遵循length constraints，然而这些方法面临两个问题：（1）低泛化性：文本类型的广泛性和constraints的特异性（比如精确的constraint: 500 words,粗粒度的constraint: 500-600words，小于500words）导致training-based 方法无法实现很好的泛化效果，我们在section 3证明了这一点。（2）inferior controllability：尽管这些方法提升了LCTG，但仍然存在不可忽视的gap。
However, the ongoing enhancements in LLM capabilities have yet to deliver the expected performance in LCTG while ensuring semantic integrity \cite{foster2024token,wang2024positionid,song2024hansel}. \citet{lift} reports that even advanced LLMs (e.g., GPT-4 Turbo \citep{gpt4}) violate the given length constraints almost 50\% of the time. To address this, training-based methods \citep{DisentDPO,lift,promptRein,ruler} have been studied to reinforce LLMs' adherence to length constraints, yet they face two key challenges: (1) \textbf{Limited generalization}: Since text types are diverse and length constraints vary widely (e.g., ranging from an exact 500 words to coarse intervals like 500-600 words or below 500 words), training-based methods often fail to generalize effectively across different settings, as demonstrated in Appendix \ref{sec:Training-based methods performance}. (2) \textbf{Inferior controllability}: These methods strengthen LCTG by enforcing implicit length modeling during generation in a top-down manner via training, lacking the decomposition and targeted enhancement of LCTG sub-capabilities, thereby limiting their progress \citep{retkowski2024zero}.

% To fill this gap，我们以人类为参考，对LCTG进行了自底向上的、解耦的子能力分解。
% 在撰写一个1000字的故事时，人类通常会首先规划各部分的内容和字数分配。而后在写作阶段，持续统计字数，并结合plan撰写文本。
% 在这个过程中，从基础到进阶依次考验了四个能力：（1）感知——正确识别一个单词作为基本单位；（2）计数——正确计数序列中单位的数量；（3）规划——合理规划各部分字数来满足length constraints；（4）生成误差——生成过程中遵循长度约束。
To fill this gap, we take humans as a reference and conduct a bottom-up decomposition of sub-capabilities for LCTG. When writing a 500-word story, humans typically begin by planning the content and word allocation for each section. During writing, they continuously track the word count and compose the text in alignment with the plan.
This process progressively tests four key abilities: (1) \textbf{Identifying} and splitting the words correctly. (2) \textbf{Counting} the words precisely. (3) \textbf{Planning} the word counts of each part to meet the length constraints. (4) \textbf{Aligning} the generated text with length constraints while ensuring semantic integrity.

On this basis, we conduct a decoupled error analysis of LCTG. The experimental results indicate that $\text{counting error} > \text{identifying error} > \text{aligning error} \gg \text{planning error}$. This suggests that deficiencies in fundamental capabilities are the primary cause of LCTG’s inferior performance. Meanwhile, it further explains why training-based approaches struggle to enhance LCTG effectively, as they are unable to provide fine-grained supervision signals for these fundamental capabilities.

Building upon this, we propose \textsc{MarkerGen}, a simple-yet-effective, plug-and-play method for achieving high-quality LCTG.
Specifically, to address LLMs' weaknesses in identifying and counting, we integrate external tokenizer and counter to track exact length information. To effectively convey these information to LLMs, we design an decaying interval insertion strategy that dynamically injects length markers during the generation process, enabling explicit length modeling while minimizing disruptions to semantic modeling.
Furthermore, to mitigate alignment issues, we propose a three-stage decoupled generation paradigm that decouples semantic constraints from length constraints, ensuring that length constraints are better met without compromising content quality.

We conduct experiments with five LLMs on five benchmarks to validate the generalizability of \textsc{MarkerGen}, covering cross-task (summarization, story generation, QA, heuristic generation), cross-scale (from 10+ to 1000+ words), cross-lingual (English and Chinese) and cross-granularity (precise and rough constraints) settings.
Experimental results demonstrate that under precise length constraints, \textsc{MarkerGen} reduces length errors by 12.57\% compared to baselines (with an average absolute error of 5.57\%), while achieving higher quality scores and incurring only 67.6\% of the cost. 
In range-based length constraints, \textsc{MarkerGen} achieves a 99\% acceptance rate, further validating its effectiveness. 
Finally, we probe into the working
mechanism of \textsc{MarkerGen} through attention analysis: shallow layers primarily handle length modeling through markers, whereas deeper layers concentrate more on semantic modeling.

% 我们通过注意力分析，揭示了在\textsc{MarkerGen}方法下LLM通过marker在浅层进行长度建模，而在深层专注语义建模的working mechanism。

% \begin{itemize}
%     \item  We conduct a \textbf{systematic bottom-up analysis} of LLMs’ length-controllable text generation, deconstructing the perception, counting, planning, and generation errors that contribute to length modeling inaccuracies.

%     \item We propose PnPLPB, a \textbf{two-stage plug-and-play framework}. It incorporates a rule-based assisted decoding length discriminator for marker insertion and a two-stage generation strategy that decouples length constraints from semantic generation to ensure both accuracy and quality.

%     \item We conduct cross-task, cross-scale, and cross-granularity experiments using Llama 3.1 and Qwen2.5 models, demonstrating that PnPLPB achieve absolute error of \textbf{5\%} avg while improving generation quality.
    
%     \item We further analyze the impact of marker insertion density and other hyperparameters on model performance, and through interpretability studies, we reveal how PnPLPB enhances LLMs' ability to precisely model text length.

% \end{itemize}

%% file: prelimilary.tex
\section{Preliminaries}
\label{sec:Preliminaries}

\begin{figure*}[t]
    \centering
    % 主图布局容器
    % \vspace{-1.0cm}
    
    % 子图排列 (使用subcaptionbox实现专业标注)
    \begin{minipage}[b]{0.46\textwidth}
        \centering        \includegraphics[ width=0.9\textwidth]{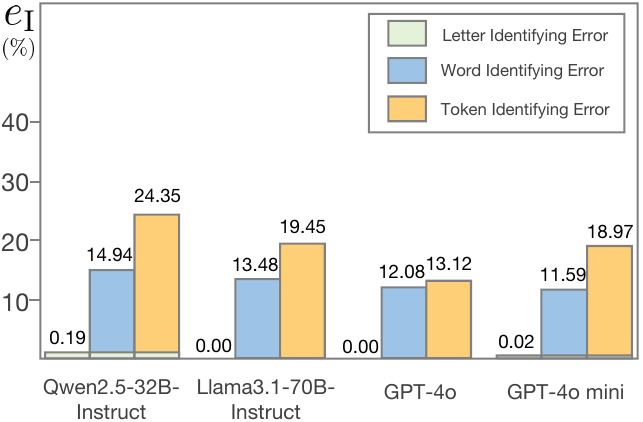}
        \subcaption{Identifying error analyses\label{fig:figure2-a}}
        \vspace*{-5pt} % 微调标注间距
    \end{minipage}
    \hfill
    \begin{minipage}[b]{0.53\textwidth}
        \centering
        \includegraphics[height=4.5cm, width=\textwidth]{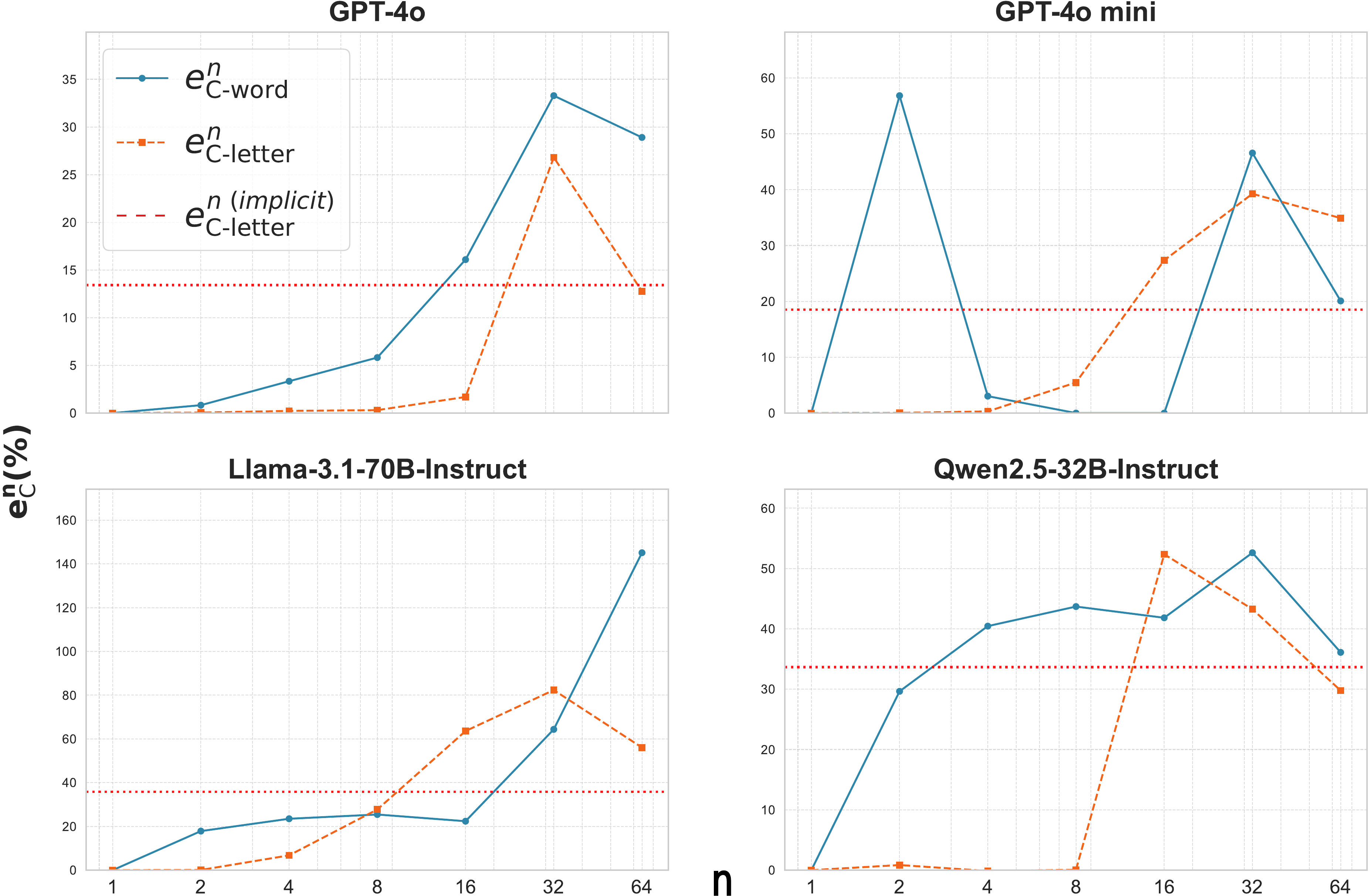}
        \subcaption{Counting error analyses\label{fig:figure2-b}}
        \vspace*{-5pt}
    \end{minipage}
    
    \caption{Error analyses of fundamental abilities in LCTG across LLMs. }
    \label{fig:figure2}
    \vspace{-3mm} % 标题间距优化
\end{figure*}

% \subsection{Decomposing LCTG Sub-Capabilities}
\label{sec:Decomposing Length Control Errors}
%我们将 LLM 的长度可控文本生成过程类比于人类在此任务中的认知机制进行建模。具体而言，模型首先根据任务需求和长度指令进行语义与长度规划。在此规划的指导下，语义空间在生成过程中以 word 级别逐步扩展，并伴随隐式的计数过程。同时，长度估计作为实时约束项，动态调控进一步的扩展。最终，模型在保证语义完整性的前提下，尽可能对齐目标长度要求。从这一视角出发，LLM 的最终生成误差可自底向上系统性地分解为识别误差（Identifying error）、计数误差（Counting error）、规划误差（Planning error） 和 对齐误差（Aligning error），它们分别影响模型对目标长度约束的对齐能力。
We model the LCTG process of LLMs by drawing an analogy to human patterns in this task. Specifically, the model first performs content and length planning based on task requirements and length constraints. Under this plan, the semantic space expands progressively at the word level during generation, accompanied by an implicit counting process. Meanwhile, length estimation acts as a real-time constraint, dynamically regulating further extension. Ultimately, the model strives to align the length constraints while preserving semantic integrity. From this perspective, the overall \textbf{LCTG} ability of LLMs can be systematically decomposed in a bottom-up manner into \textbf{Identifying}, \textbf{Counting}, \textbf{Planning}, and \textbf{Aligning} sub-capabilities (Figure \ref{fig:Figure0}). 
% 以下我们通过error分析来探究LLMs这些子能力的掌握情况。
Below we explore LLMs' mastery of these abilities through detailed error 
analysis on TruthfulQA dataset \citep{TruthfulQA}.

\subsection{Identifying Error}
\label{sec:identifying_error}
%识别误差指的是模型在判断基本长度单位（如单词）时的错误识别，导致其估算的文本长度与实际长度之间存在偏差。为系统性分析该误差，我们设计了一项实验，要求模型逐个识别并累加长度单位，并将其预测的计数结果与真实值进行比较。实验结果验证，在逐一累加的设定下，计数误差不会发生，这意味着最终的长度误差完全来源于识别误差。

% To systematically analyze this error, we instruct the model to recognize the length units of given text one by one in format:  
% Identifying error refers to the misidentification of fundamental length units (e.g., words), leading to discrepancies between the model’s estimated and actual text length. 
% To systematically analyze this error, we design an experiment where the model is prompted to recognize and accumulate length units sequentially and compare its predicted count against the ground truth. Experimental results confirm that in the one-by-one accumulation setting, counting errors do not occur, meaning that the final length error entirely stems from identifying error. The identifying error \( e_2 \) is computed as:

Identifying error refers to the misidentification of fundamental length units (e.g., words), leading to discrepancies between the model’s estimated and actual text length.
To systematically analyze this error, we instruct the model to recognize the length units of given text one by one. If we define a word as the length unit, the model should output like: ``\texttt{The [1 word] quick [2 words] fox [3 words] ...}''.
On this basis, we calculate the identifying error rate $e_{\text{I}}$ as follows:
\begin{equation}
e_{\text{I}} = \frac{|N_{\mathrm{pred}}^{\mathrm{1}} - N_{\mathrm{true}}|}{N_{\mathrm{true}}} 
\label{eq:e1}
\end{equation}
% 其中，\( N_{\mathrm{pred}}^{\mathrm{1p}} \) 是模型在逐个预测（one-by-one）设置下的预测计数，\( N_{\mathrm{true}} \) 是实际计数。
where \( N_{\mathrm{pred}}^{\mathrm{1}} \) is the model’s predicted final count with 1 as count interval, and \( N_{\mathrm{true}} \) is the actual count.
% 如图 \ref{fig:figure2} 所示，我们在 TruthfulQA 数据集上进行错误识别实验，分别在词（word）、子词（token）和字母（letter）级别上进行评估，其中参考文本作为计数依据。为了进行受控的消融研究，我们定义字母级计数，使得计数的字母数与参考文本的词数相匹配。实验结果表明，大规模语言模型（LLMs）在识别过程中存在显著错误，并且其误差幅度呈现出明确的模式：\textit{字母 $\ll$ 词 $<$ 子词}。总体而言，我们的发现如下：
% \vspace{5pt}  
% \noindent\textbf{发现 1：} LLMs 存在显著的识别误差，这是其长度建模不准确的根本原因。  
% \vspace{3pt}  
% \noindent\textbf{发现 2：} 语言模型感知长度的基本单位是词（word），这表明 LLMs 主要通过\textbf{语义感知}（semantic perception）建模长度，而非\textbf{解码机制}（decoding mechanics）。
% 
% 我们将$e_1$减去把每个word替换成letter ``\texttt{A}''时得到的error rate(此时并不考验identifying能力)，作为修正后的$e_1$。
We subtract the error rate obtained when replacing each word with the letter ``\texttt{A}'' (which barely assess the identifying ability) from $e_{\text{I}}$ to further eliminate the influence of other potential factors.
We explore the word and token as length unit respectively,
as shown in Figure~\ref{fig:figure2-a}.

% the results indicate that LLMs exhibit substantial identifying errors, with the magnitude following a clear pattern: \textit{letter $\ll$ word $<$ token}. Overall, we find that:
% \vspace{2pt}  
\begin{finding}
\label{find1}
LLMs exhibit notable $e_{\text{I}}$ with both word and token as unit, showcasing their deficiencies in fundamental identifying ability.
\end{finding}
\begin{finding}
\label{find2}
Word yields lower $e_{\text{I}}$ than token, indicating that LLMs conduct length modeling primarily based on \textbf{semantic perception} rather than \textbf{decoding mechanics}.
\end{finding}
% \begin{mdframed}
% \noindent\textbf{Finding 1:} LLMs exhibit notable error rates under both settings, showcasing their deficiencies in fundamental identifying ability.

% \noindent\textbf{Finding 2:}  Word yields lower $e_{\text{I}}$ than token, indicating that LLMs model length primarily based on \textbf{semantic perception} rather than \textbf{decoding mechanics}.  
% \end{mdframed}

% \vspace{1pt}  
% \noindent\textbf{Finding 2:} Word yields lower $e_{\text{I}}$ than token, indicating that LLMs model length primarily based on \textbf{semantic perception} rather than \textbf{decoding mechanics}.  

\begin{figure*}[t]
    % 下面两张图并排
    \begin{minipage}{0.51\textwidth}
        \centering
        \includegraphics[height=4.8cm]{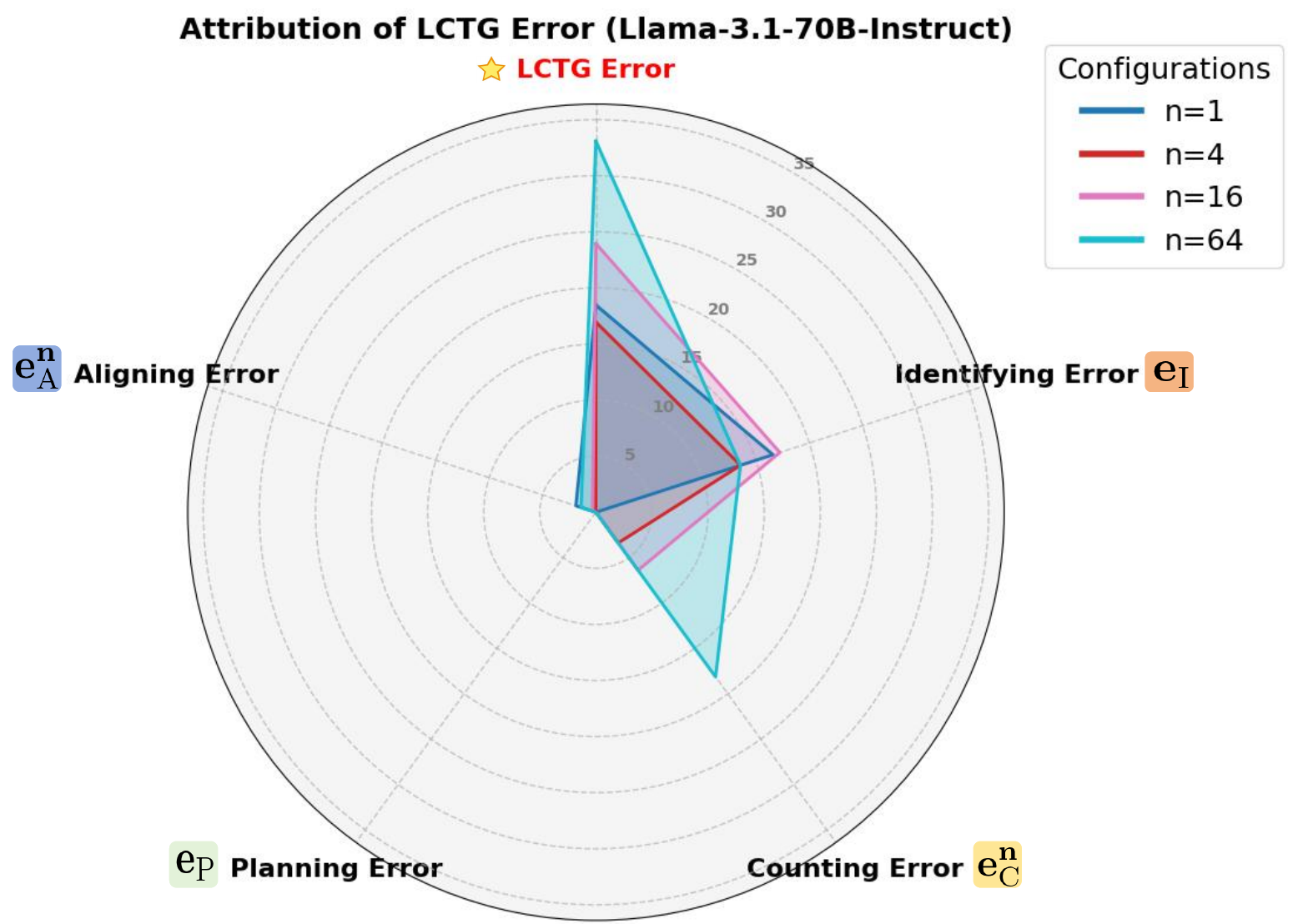}
    \end{minipage}
    \hfill
    \begin{minipage}{0.51\textwidth}
        \centering
        \includegraphics[height=4.8cm]{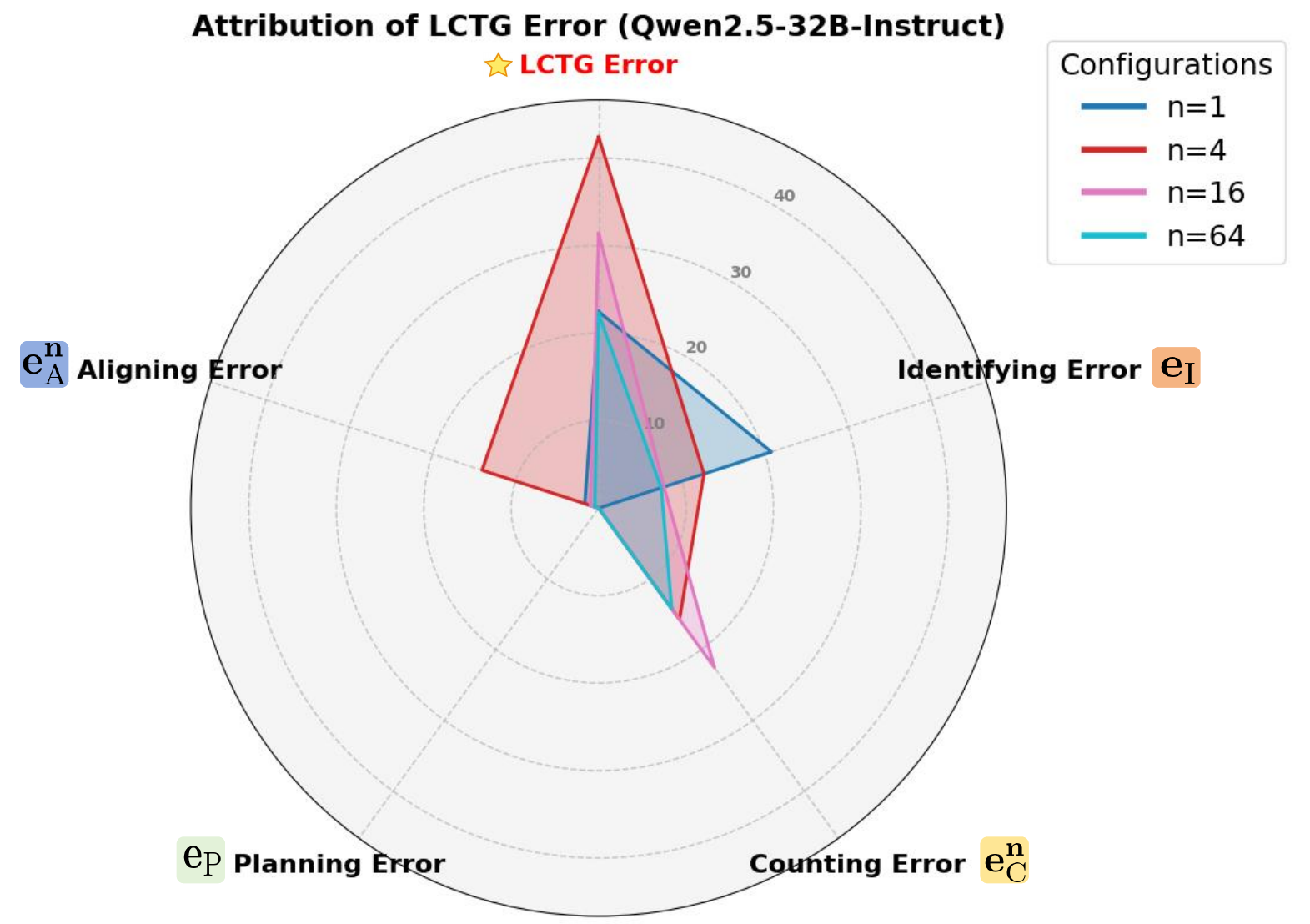}
    \end{minipage}

    \caption{Absolute contribution of LCTG sub-capability deficiency on overall LCTG error across LLMs.}
    \vspace{-12pt}
    \label{fig:Figure3}
\end{figure*}

\subsection{Counting Error}
\label{sec:counting_error}
% 计数误差（Counting error）指的是对给定序列中元素的不准确枚举，导致其长度偏离预期。我们通过引导 LLMs 计算按 \( n \) 个元素分组的序列来分析这种误差，其中 \( n = 1, 2, 4, 8, \dots \)。当 \( n = 1 \) 时，对应于识别误差（Identifying error）（详见 \ref{sec:identifying_error}）。由于 LLMs 在词感知上表现出识别误差，我们将观察到的计数误差 \( e_{23} \) 分解为识别误差 \( e_2 \)（见公式 \ref{eq:e2}）和纯计数误差 \( e_3 \)，其表达式如下：
% , where \( n = 1, 2, 4, 8, \dots \)
% n越大，越考验counting能力
Counting error refers to the inaccurate enumeration of length units in a given sequence, leading to deviations from the intended length. 
We analyze this error by prompting LLMs to count sequences with varied interval $n$. 
The case of \( n = 1 \) corresponds to identifying error (see \S\ref{sec:identifying_error}). 
A larger $n$ poses a greater challenge for counting accuracy.
To decompose counting error from identifying error, we calculate $e_{\text{C}}^n$ as follows:
\begin{equation}
e_{\text{IC}}^n = \frac{|N_{\mathrm{pred}}^{n} - N_{\mathrm{true}}|}{N_{\mathrm{true}}} 
\label{eq:e12}
\end{equation}
\begin{equation}
e_{\text{C}}^n = e_{\text{IC}}^n - e_{\text{I}}
\label{eq:e2}
\end{equation}

% 如图 \ref{fig:figure2} 所示，\ref{sec:identifying_error} 中的研究结果证实，LLMs 在字母级别上的识别误差可以忽略不计，因此字母级的计数误差可以作为衡量纯计数能力的直接指标。我们还引入了一个基线（letters），其中 LLM 在未被明确要求一次计数 \( n \) 个元素的情况下执行隐式计数。基于实验结果和观察到的模式，我们得出以下关键发现：
% \vspace{5pt}  
% \noindent\textbf{发现 3：} **LLMs 存在计数误差，并且误差随着 \( n \) 的增大而增加。** 无论是计数单词（\( e_2 + e_3 \)）还是计数字母（\( e_3 \)），当 \( n \) 较大时，对计数精度的要求更高，导致误差迅速增加。
% \vspace{3pt}  
% \noindent\textbf{发现 4：} **显式计数结合更精细的计数间隔可以提高 LLMs 的长度建模精度。** 在较小的 \( n \) 下，$\mathbf{e_{\mathit{letters}}}$ 明显低于 $\mathbf{e_{\mathit{letters}}}$（基线），表明通过较小单位的逐步显式计数有助于减少隐式长度建模中的误差。
% As shown in Figure \ref{fig:figure2}, 
Since LLMs exhibit negligible identifying error at the letter level (Figure~\ref{fig:figure2-a}), error of counting letter serves as a direct measure of pure counting ability. We also include a commonly used baseline where the LLMs conduct implicit counting (directly output the length of the entire given text). The results are shown in Figure~\ref{fig:figure2-b}.
% direct count without being explicitly prompted to count \( n \) items at a time. Based on the experimental results and observed patterns, we derive the following key findings:

\begin{finding}
\label{find3}
% Naive的隐式地counting会导致明显的error
\textbf{Naive implicit counting can lead to significant errors.} 
\end{finding}
% Whether counting words (\( e_1 + e_2 \)) or letters (\( e_2 \)), the higher demands on counting ability at larger \( n \) lead to a rapid increase in errors.

\begin{finding}
\label{find4}
\textbf{Explicit counting combined with fine-grained intervals leads to better length modeling.}  At smaller \( n \), the error of explicit counting is significantly lower than that of implicit counting.
\end{finding}

\begin{table}[t]
  \centering
   \renewcommand\arraystretch{1.3}
  \setlength{\tabcolsep}{0.3em} 
  \small
  % \resizebox{\linewidth}{!}{%
    \begin{tabular}{lcccc}
    \toprule
      \textbf{Models} & $e_{\text{P}}$  & $s_{\text{P}}$ & $\Delta E$ ($\downarrow$) & $\Delta S$ ($\uparrow$) \\
      \midrule
      % \hline
      GPT-4o                 & 0.06 & 4.28 & -5.31  & 0.05 \\
      GPT-4o mini           & 0.33 & 3.90 & +2.11  & 0.03 \\
      Llama-3.1-70B-Instruct & 0.00 & 3.90 & -0.63 & 0.04 \\
      Qwen2.5-32B-Instruct   & 0.04 & 4.22 & -8.93  & 0.02 \\
      \bottomrule
    \end{tabular}  \caption{\label{tab:planning_error}$e_\text{P}$ and $s_\text{P}$ denote planning error and planning quality score of LLMs. $\Delta E$ and $\Delta S$ quantify the LCTG error reduction and text quality gain from two-stage generation over one-stage generation.}
    \vspace{-10pt}
\end{table}

\vspace{-2pt}

\subsection{Planning Error}
\vspace{-2pt}
\label{sec:planning_error}
%规划误差指的是在长度约束下，不同文本部分的字数分配不当，导致最终生成的总长度未能满足目标要求。为研究该误差，我们计算各部分规划长度之和与目标长度之间的偏差，即e1:

% Planning error refers to the misallocation of word counts across different sections of the text under a given length constraint, resulting in a total length that deviates from the target requirement. To quantify this error, we measure the discrepancy between the sum of the planned segment lengths and the target length, defined as:

% \begin{equation}
% e_{\text{P}}= \frac{|N_{\mathrm{plan}} - N_{\mathrm{target}}|}{N_{\mathrm{target}}} 
% \label{eq:e3}
% \end{equation}

% where \( N_{\mathrm{plan}} \) denotes the total word count allocated by the model, and \( N_{\mathrm{target}} \) represents the specified target length.
Planning error refers to the misallocation of word counts across different sections, leading to a discrepancy from target length. For given query and precise length constraint \( N_{\mathrm{target}} \), 
we prompt LLMs to explicitly plan both content and length for each part of the response.
% we prompt LLMs to generate explicit planning statements regarding their responses, including segment-wise length allocations and content structuring. 
We assess the quality \footnote{We use Qwen-Plus \citep{qwen25} as the judge with a scoring range of [1, 5]. See corresponding prompts in Appendix \ref{sec:appen_planning}} of the plan $s_{\text{P}}$, and calculate the planning error rate $e_{\text{P}}$ as:
\begin{equation}
e_{\text{P}}= \frac{|N_{\mathrm{plan}} - N_{\mathrm{target}}|}{N_{\mathrm{target}}} 
\label{eq:e3}
\end{equation}
% We additionally assess the quality \footnote{All quality evaluations use Qwen-Plus \citep{qwen25} as the judge with a scoring range of [1, 5]. See corresponding prompts in Appendix 1.} of the plan $s_{\text{P}}$.
% Additionally, we compare the effects of \textbf{planning followed by generation} versus \textbf{direct generation} on both length accuracy and content quality. 
where \( N_{\mathrm{plan}} \) denotes the total word count allocated by the model.
Meanwhile, we calculate the reduction in final length error (\(\Delta E\)) and the improvement in content quality (\(\Delta S\)) achieved by \textbf{planning followed by generation} compared to \textbf{direct generation}.
The results are shown in Table~\ref{tab:planning_error}.
% Through these analyses, we derive two key findings:

\begin{finding}
\label{find5}
\textbf{LLMs exhibit strong planning ability.} 
The generated plan effectively meets the length constraints while achieving a quality score of around 4, demonstrating well-structured content allocation.
\end{finding}
\begin{finding}
\label{find6}
\textbf{Planning before generation brings better results.} Compared to direct generation, executing planning and generating sequentially for decomposition reduces length deviations while enhancing semantic quality.
\end{finding}

\subsection{Aligning Error}  
\label{sec:aligning_error}
\begin{figure}[ht]
    %\vspace{-1.0cm}
  \includegraphics[width=\columnwidth,height=4cm]{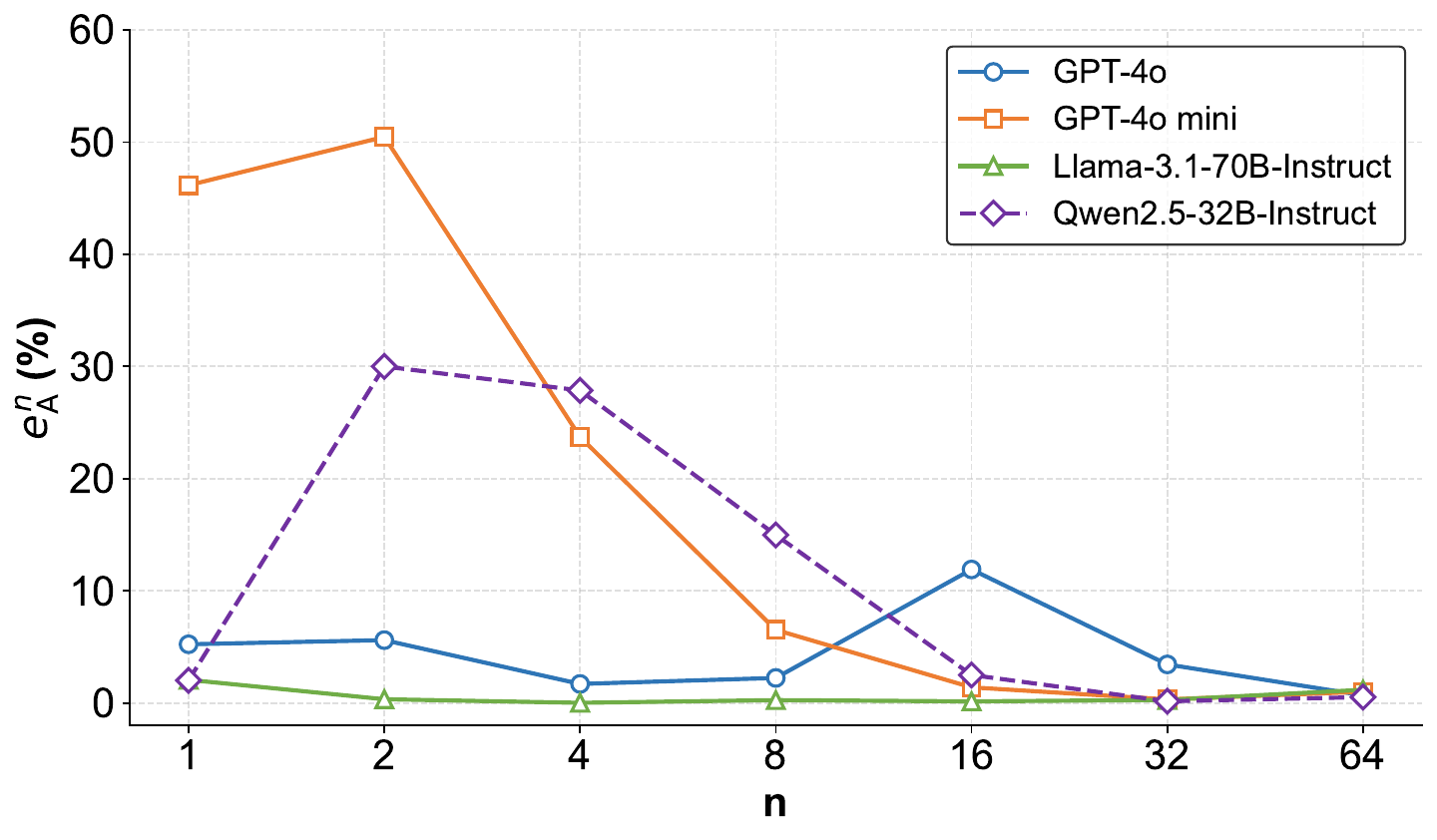}
  \vspace{-20pt}
  \caption{Aligning Error across varied length intervals.}
  \vspace{-12pt}
  \label{fig:Figure4}
\end{figure}

Aligning error refers to the discrepancy between the model’s perceived length and the target length, arising from the challenge of maintaining semantic integrity while adhering to length constraints. We calculate aligning error as follows:  
\begin{equation}
e^n_{\text{A}} = \frac{|N^n_{\mathrm{pred}} - N_{\mathrm{target}}|}{N_{\mathrm{target}}} 
\label{eq:e4}
\end{equation}
where \( N^n_{\mathrm{pred}} \) represents the model’s perceived length with counting interval $n$, i.e., the length the model assumes it has generated. 
We calculate and show the $e_A^{n}$ in Figure \ref{fig:Figure4}.
% 
% , which is affected by identifying and counting errors.

\begin{finding}
\label{find7}
\textbf{Smaller counting intervals introduce greater aligning error.}
By closely analyzing cases, we find that frequent explicit counting interferes with semantic modeling, causing early termination of generation and poor alignment. In contrast, larger length intervals approximate implicit counting, preserving a more natural generation process.
\end{finding}

\vspace{-4pt}
\subsection{LCTG Error}  
LCTG error refers to the discrepancy between the actual length of generated text and the target length:
\begin{equation}
E = \frac{|N_{\mathrm{true}} - N_{\mathrm{target}}|}{N_{\mathrm{target}}} 
\label{eq:E}
\end{equation}
As established above, this error is systematically composed of four components: \textbf{Identifying Error} (\S\ref{sec:identifying_error}), \textbf{Counting Error} (\S\ref{sec:counting_error}), \textbf{Planning Error} (\S\ref{sec:planning_error}), and \textbf{Aligning Error} (\S\ref{sec:aligning_error}). To investigate the key factors influencing LLMs' LCTG error, we calculate their absolute contributions $\dot{e}_i^{n}$ under different length interval \( n \) as follows:  
\begin{equation}
\dot{e}_i^{n} = \frac{e_i^{n}}{e_\text{I} + e_\text{C}^{n} + e_\text{P} + e^n_\text{A}} \times E^{n} ,\ \ i\in[\text{I,C,P,A}]
\label{eq:ei}
\end{equation}
% \begin{equation}
% e_i^{n} = \frac{e_i}{e_1 + e_2 + e_3 + e_4} \times E^{n} 
% \end{equation}
The results are shown in Figure~\ref{fig:Figure3}. Further details can be found in \ref{sec:Sub-ability Decomposition}. 
% presents an error proportion analysis of Llama-3.1-70B-Instruct and Qwen2.5-32B-Instruct across four representative settings. Figure \ref{fig:Figure4} illustrates the proportion of early-stopped items caused by Aligning Error as \( n \) varies across four LLMs. Further details can be found in \ref{sec:appendix}. Based on these results, we summarize the following key findings:

\begin{finding}
\label{find8}
\textbf{LCTG error is primarily attributed to fundamental deficiencies in length modeling, following the order of}  
 \textbf{Counting Error} $>$ \textbf{Identifying Error} $>$ \textbf{Aligning Error} $\gg$ \textbf{Planning Error}.  
Thus, as counting interval increases, the accumulation of counting errors leads to a corresponding rise in LCTG error.
\end{finding}

 \begin{figure*}[t]
    \centering 
\includegraphics[width=0.95\textwidth]{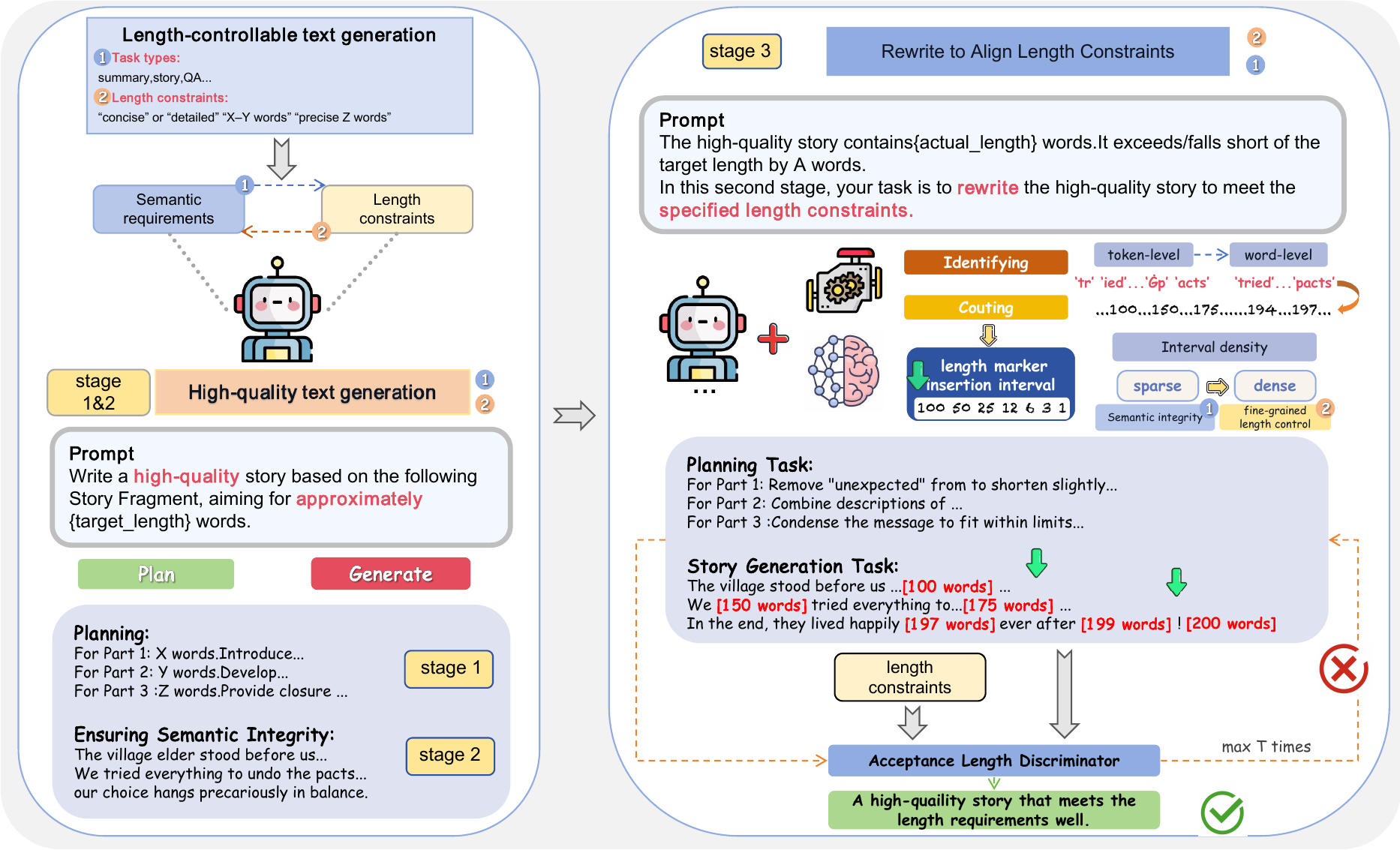}
\vspace{-5pt}
  \caption{Overview of \textsc{MarkerGen}.}
    \vspace{-12pt}
  \label{fig:Figure5}
\end{figure*}

% \vspace{-5pt}  
% \subsection{LLM Decoding Mechanism}
% LLMs generate text in an autoregressive manner, predicting each token sequentially based on previously generated tokens. Given an input sequence $X = \{x_1, x_2, ..., x_n\}$, the generation process follows:

% \begin{equation}
%     x_{t+1} \sim P(X | x_{\leq t}),
% \end{equation}

% where each token $x_{t+1}$ is sampled from the model’s probability distribution conditioned on prior context. This iterative decoding continues until a predefined stopping criterion is met, such as reaching a maximum length or generating an end-of-sequence (EOS) token.

%% file: method.tex
\section{Methodology}

Based on the analyses and findings above, we propose \textsc{MarkerGen}, a simple-yet-effective plug-and-play method to help LLMs attain better LCTG performance, as shown in Figure~\ref{fig:Figure5}. This method consists of two key modules: (1) \textbf{Auxiliary Marker Insertion Decoding} mechanism, which explicitly enhances length modeling during generation; (2) \textbf{Three-Stage Decoupled Generation} scheme, which decouples length constraints from semantic content generation to further improve LCTG performance.

\subsection{Auxiliary Marker Insertion Decoding}
\label{sec:Auxiliary Length Marker Insertion Decoding}

\paragraph{External Tool Invocation.} 
Our analysis in \S\ref{sec:Preliminaries} reveals that LLMs exhibit significant identifying and counting errors, which directly contribute to inaccuracies in length modeling.
To mitigate these fundamental deficiencies, we introduce external tokenizer and counter for unit recognition and counting, respectively. 
As Finding \ref{find1} indicates that LLMs perceive words better than tokens, we select words as the length unit.
% Theoretically, these errors can be completely removed.
% 为了弥补这两个基础能力，我们考虑引入external tokenizer and counter来分别识别word和计数。理论上，这两个error可以被完全消除。
% 
\paragraph{Length Information Injection.}
With precise length information, we consider feeding it into the model for length modeling. 
Since Finding \ref{find3} indicates that LLMs' inherent implicit length modeling leads to significant errors and is inconvenient for incorporating external length information, we actively insert precise length markers during generation to enable explicit length modeling:
\begin{equation}
\small
\begin{gathered}
    \text{Len}(x) = \text{Counter}(\text{Tokenizer}(x))\\
    x_{t+1} = \begin{cases}
        \text{Marker}(\text{Len}(x_{\leq t})),
         & \text{if} \  \mathcal{S}(\text{Len}(x_{\leq t}),N)   \\
         \text{Sampling}(P(x_{t+1} | x_{\leq t})),
        & \text{else}
    \end{cases}
\end{gathered}
\end{equation}
where $P(x_{t+1} | x_{\leq t})$ is the LLM's probability distribution for next token, $\text{Marker}$ defines the marker format (e.g., [20 words], we discuss the effects of varied marker formats in Appendix \ref{sec:LM_form}), $\mathcal{S}$ is the strategy that determines whether to insert a marker based on current length $\text{Len}(x)$ and target length $N$.
By treating the inserted markers as anchors, LLMs can continuously adjust the expected length of content  to be generated during the generation process, thereby reducing the final LCTG error.
% 得到这些准确的长度信息后，我们需要将其传送给模型帮他进行长度建模。模型原本是自己在隐空间进行长度建模，我们考虑在生成的过程中，间隔地插入marker来帮助模型进行显示长度建模。不同于第二节模型自行插入的是，此时maker插入的时机和所显示的长度信息由外部工具决定，因此是

\paragraph{Decaying Interval Marker Insertion Strategy.}
% 最naive的insertion strategy 是等间隔地插入marker，我们记作$\mathcal{S}_$
The most naive insertion strategy involves placing markers at uniform intervals, which we denote as $\mathcal{S}_{uni}$.
% 然而，根据Finding 4和5说明，较小的insertion interval $n$ 会带来较好的长度建模，但会破坏semantic modeling，而$n$较大时情况相反。
However, according to Findings \ref{find4} and \ref{find7}, a smaller insertion interval $n$ improves length modeling but compromises semantic modeling, whereas a larger $n$ exhibits the opposite effect.
Considering this, we propose a strategy $\mathcal{S}_{dec}$, where $n$ decays exponentially during the generation process:
\begin{equation}
    \small
    \mathcal{S}_{dec}(x,N)= \begin{cases}
        \text{True}, & \text{if} \ x \in\{N-\text{int}(2^{-i}\times N)\}_{i\in \mathbb{N}}   \\
        \text{False},& \text{else}
    \end{cases}
\end{equation}
Taking $N=200$ as an example, the maker will be inserted behind the 100th, 150th, 175th, ... words.
% 这样，在前期模型可以专注于semantic modeling。随着生成的进行，模型越来越关注对长度的控制，最终导向较小的LCTG误差。因此$\mathcal{S}_{dec}$能较好地平衡semantic 和 length modeling。
At the early stage of generation, the model primarily focuses on semantic modeling. As the generation progresses, it increasingly emphasizes length control, ultimately leading to a smaller LCTG error. 
% Considering that aligning error stems from semantic modeling 被干扰，
Consequently, $\mathcal{S}_{dec}$ effectively balances semantic modeling and length modeling.

\vspace{-5pt}
\subsection{Three-Stage Decoupled Generation}
\label{sec:Two-Stage High-Quality Text Generation}
% Finding 6已经证明了Planning before generation的两阶段方法相比于直接生成能导向更好的LCTG results。
% Finding 7证实了aligning error主要源自于语义建模被影响导致的生成过程早停止问题
% 为了减小aligning error同时增强文本quality，我们将原有的planning before generation scheme进一步结耦为三阶段。
% According to Finding 6, the two-step method of planning before generation yields improved LCTG results compared to generating directly. 
% 为了
Finding \ref{find7} validates that aligning error primarily arises from the inferior semantic modeling, which causes premature termination of the generation process.
% the original planning before generation scheme虽然通过将plan进行结耦取得了比direct generation更好的效果（Finding 6），但长度建模过程和语义建模过程仍然耦合。
% 为此，我们提出Three-Stage Decoupled Generation scheme，to further reduce aligning error and improve text quality。
% To reduce aligning error and improve text quality, we further decouple the original planning before generation scheme into three distinct stages.
%
While the planning before generation scheme alleviates interference in semantic modeling by decoupling the planning process (Finding \ref{find6}), it still entangles length modeling with semantic modeling.
To mitigate this, we introduce a three-stage decoupled generation scheme to further reduce the alignment error and improve the text quality, as illustrated in Figure~\ref{fig:Figure5}.
\paragraph{Stage One: Planning.} The model generates a reasonable plan based on the input query and length constraints, including the content of each section and the word allocation.
% 模型根据plan中的内容规划进行高质量的语义建模和生成，without enforcing strict length constraints.
\paragraph{Stage Two: Ensuring Semantic Integrity.} The model focuses on semantic modeling to generate a high-quality response per the plan without being strictly required to adhere to length constraints.
\paragraph{Stage Three: Aligning Length Constraints.} Responses generated in stage two are usually of high quality but may not meet length restrictions. To refine them, we use these non-compliant responses as input and apply the Auxiliary Marker Insertion Decoding mechanism for rewriting. The \textbf{rewriting requirements} include: (1) Retaining the high-quality semantic modeling of the input content. (2) Strictly adhering to the specified length constraints.
In terms of \textbf{workflow}, the model is required to: (1) Firstly analyze the previous stage's response for potential improvements; (2) If its output does not meet the length constraints, it will be regenerated up to $T$ times or until the constraints are met.

% A single-stage approach requires the model to simultaneously satisfy both length constraints and semantic quality, making the task highly challenging. We examine two variants:  
% (1) \textbf{Implicit}, where the model relies on implicit length control without explicit markers, adjusting length based on learned patterns.  
% (2) \textbf{Insert}, which enhances length awareness by incorporating explicit length markers during decoding (Section \ref{sec:Auxiliary Length Marker Insertion Decoding}).

% \paragraph{Two-Stage}  
% To reduce the difficulty of simultaneous optimization, we adopt a two-stage strategy to decouple length control from semantic generation. In the first stage, the model generates text with a focus on semantic quality, without enforcing strict length constraints. In the second stage, non-conforming outputs undergo controlled rewriting using the auxiliary length marker insertion strategy to ensure precise length alignment while preserving coherence.
% Our default setting for MAKERGEN incorporates the Three-Stage Decoupled Generation along with the adaptive auxiliary length marker insertion strategy, as illustrated in Figure~\ref{fig:Figure5}.
% 通过三阶段的
See Appendix ~\ref{sec:prompt-three-stage}  for prompts of each stage.

%% file: exp.tex
\section{Experiments}
We conduct comprehensive experiments to examine \textsc{MarkerGen}. Specifically, we validate its effectiveness in \S\ref{sec:performance}, analyze its generalizability in \S\ref{sec:Generalizability}, explore the impact of its key components in \S\ref{sec:ablation}, 
and provide further insights into its mechanism in \S\ref{sec:analysis}.
Hyperparameter choices and additional analyses are provided in Appendix~\ref{sec:exp details}.

\begin{table}[t]
    \renewcommand\arraystretch{1.1}
    \small
    \centering
    \setlength{\tabcolsep}{0.20em} 
    \begin{tabular}{lcc}
    \toprule
    Benchmarks&Ability Tested&Length (words)\\
     \midrule
     CNN/DailyMail&\multirow{2}{*}{Summarization} & \multirow{2}{*}{18-165}\\
     \citep{CNN/DailyMail}&&\\
    \cdashline{1-3}
    HANNA&\multirow{2}{*}{Story Generation} & \multirow{2}{*}{139-995}\\
     \citep{HANNA}&&\\
     \cdashline{1-3}
     TruthfulQA&Question & \multirow{2}{*}{101-294}\\
     \citep{TruthfulQA}&Answering&\\
     \cdashline{1-3}
    HelloBench&Heuristic LCTG\&& \multirow{2}{*}{489-1450}\\
     \citep{HelloBench}&Open-ended QA&\\
     \cdashline{1-3}
    GAOKAO&History& \multirow{2}{*}{71-901}\\
     \citep{gaokao}&Open-ended QA&\\
    \bottomrule
    \end{tabular}
    \vspace{-0.2cm}
    \caption{Benchmarks Introduction.}
    \label{tab: benchmark}
    \vspace{-0.5cm}
\end{table}

\subsection{Experimental Settings}

\begin{table*}[t]
  \centering
  \small
    \setlength{\tabcolsep}{0.40em} 
    \renewcommand{\arraystretch}{1.3}
    \footnotesize
  \begin{tabular}{l c *{12}{cc} c}
    \toprule
    \multirow{4}{*}{\textbf{Benchmarks}} & \multirow{4}{*}{\textbf{Methods}} 
    & \multicolumn{6}{c}{\textbf{Qwen2.5 Series}} & \multicolumn{4}{c}{\textbf{Llama3.1 Series}} & \multirow{4}{*}{\textbf{Costs}} \\
    \cmidrule(lr){3-8} \cmidrule(lr){9-12}
    & & \multicolumn{2}{c}{7B} & \multicolumn{2}{c}{14B} & \multicolumn{2}{c}{32B} & \multicolumn{2}{c}{8B} & \multicolumn{2}{c}{70B}  & \\
    \cmidrule(lr){3-4} \cmidrule(lr){5-6} \cmidrule(lr){7-8} \cmidrule(lr){9-10} \cmidrule(lr){11-12}
    & & $E$ ($\downarrow$) & $S$ ($\uparrow$) & $E$ ($\downarrow$) & $S$ ($\uparrow$) & $E$ ($\downarrow$) & $S$ ($\uparrow$) & $E$ ($\downarrow$) & $S$ ($\uparrow$) & $E$ ($\downarrow$) & $S$ ($\uparrow$) & \\
    \midrule
    \multirow{2}{*}{CNN/DailyMail} 
      & Implicit    & 30.31 & 3.04 & 12.54 & 3.15 & 11.05 & 3.21 & 15.12 & 3.04 & 11.07 & 3.09  & $1.30 \times \delta$\\
      & \textsc{MarkerGen}  & \textbf{9.92} & \textbf{3.07} & \textbf{6.06} & \textbf{3.16} & \textbf{4.82} & \textbf{3.25} & \textbf{3.36} & \textbf{3.18} & \textbf{3.18} & \textbf{3.36} & $\delta$ \\
    \cdashline{1-13}
    \multirow{3}{*}{HANNA} 
      & Implicit  & 28.55 & 3.47 & 14.86 & 3.55 & 12.03 & 3.67 & 16.68 & 3.54 & 10.44 & 3.61  & $2.37 \times \delta$\\
      & \textsc{MarkerGen}  & \textbf{8.49} & \textbf{3.50} & \textbf{5.22} & 3.55  & \textbf{3.57} & \textbf{3.72} & \textbf{2.98} & \textbf{3.60} & \textbf{2.58} & \textbf{3.63} & $\delta$ \\
    \cdashline{1-13}
    \multirow{2}{*}{TruthfulQA}
      & \multirow{1}{*}{Implicit}    & 16.7 & 4.29 & 17.9 & 4.44 & 8.7 & 4.45 & 7.21 & 4.22 & 7.64 & 4.46 & $1.75 \times \delta$\\
      & \textsc{MarkerGen}    & \textbf{9.08} & \textbf{4.33} & \textbf{7.59} & 4.43 & \textbf{4.48} & \textbf{4.54} & \textbf{3.82} & \textbf{4.25} & \textbf{2.80} & \textbf{4.48}  & $\delta$ \\
    \cdashline{1-13}
    \multirow{2}{*}{Heuristic Generation}
      & \multirow{1}{*}{Implicit}    & 35.69 & 3.42 & 21.34 & 3.80 & 12.02 & 3.80 & 21.91 & 3.72 & 27.89 & 3.74 & $1.06 \times \delta$ \\
      & \textsc{MarkerGen}& \textbf{8.51} & \textbf{4.13} & \textbf{6.35} & \textbf{4.00} & \textbf{5.34} & \textbf{4.14} & \textbf{6.03} & \textbf{4.03} & \textbf{5.03} & \textbf{3.98} & $\delta$ \\
    \bottomrule
  \end{tabular}%
  \vspace{-2pt}
\caption{\label{tab:Overall}Overall Performance of \textsc{MarkerGen} on Various Benchmarks. $E$ denotes LCTG error rate (\%) and $S$ denotes the text quality ([1, 5]) given by LLM judge. $\delta$ denotes the token cost of \textsc{MarkerGen} under each setting.}
\end{table*}

\begin{table*}[t]
  \centering
  \small
    \setlength{\tabcolsep}{0.6em} 
  \renewcommand{\arraystretch}{1}
 \footnotesize
  \begin{tabular}{ll *{8}{cc} c}
    \toprule
    \multirow{4}{*}{\textbf{Model}} & \multirow{4}{*}{\textbf{Methods}} & \multicolumn{8}{c}{\textbf{Target Length Scales}} &\multirow{4}{*}{\textbf{Costs}} \\
    \cmidrule(lr){3-10}
    & & \multicolumn{2}{c}{100} & \multicolumn{2}{c}{200} & \multicolumn{2}{c}{300} & \multicolumn{2}{c}{400} &\\
    \cmidrule(lr){3-4} \cmidrule(lr){5-6} \cmidrule(lr){7-8} \cmidrule(lr){9-10} 
     & & $E$ ($\downarrow$) & $S$ ($\uparrow$) & $E$ ($\downarrow$)& $S$ ($\uparrow$) & $E$ ($\downarrow$)& $S$ ($\uparrow$) & $E$ ($\downarrow$)& $S$ ($\uparrow$) &\\
    \midrule
    \multirow{2}{*}{\textbf{Qwen2.5-7B-Instruct}} 
    & \multirow{1}{*}{Implicit} 
      & 30.97 & 3.45 & 22.91  & 3.53 & 26.12 &3.28  &29.63 &3.08  & $1.26 \times \delta$ \\
    % \cmidrule(lr){3-11}
    & \multirow{1}{*}{\textsc{MarkerGen}} 
      & 8.26  &3.92  &9.06  &4.00  &7.67  &3.75 &5.10   &3.55  & $\delta$  \\
    \midrule
    \multirow{4}{*}{\textbf{Model}} & \multirow{4}{*}{\textbf{Methods}} & \multicolumn{8}{c}{\textbf{Length Constraint Types}} & \multirow{4}{*}{\textbf{Costs}}\\
    \cmidrule(lr){3-10}
    & & \multicolumn{2}{c}{<100} & \multicolumn{2}{c}{100-150}  & \multicolumn{2}{c}{160-200} & \multicolumn{2}{c}{>500}\\
    \cmidrule(lr){3-4} \cmidrule(lr){5-6} \cmidrule(lr){7-8} \cmidrule(lr){9-10} 
    & & $E_r$ ($\downarrow$)& $S$ ($\uparrow$) & $E_r$ ($\downarrow$)& $S$ ($\uparrow$) & $E_r$ ($\downarrow$)& $S$ ($\uparrow$) & $E_r$ ($\downarrow$)& $S$ ($\uparrow$) \\
    \midrule
    \multirow{2}{*}{\textbf{Qwen2.5-7B-Instruct}} 
    & \multirow{1}{*}{Implicit} 
      & 7.50  &3.47  & 63.00 & 4.03 & 66.00  & 4.06  & 29.50 & 2.65 & $1.07\times \delta$   \\
    % \cmidrule(lr){3-11}
    & \multirow{1}{*}{\textsc{MarkerGen}} 
      & 0.00   & 3.94  & 0.50  & 4.50 & 3.00  &4.53  & 0.00  &3.13  & $\delta$  \\
    \bottomrule
  \end{tabular}%
\caption{\label{tab:openqa}Experiments with varied length scales and constraint types on Open-ended QA subset of HelloBench.}
\vspace{-10pt}
\end{table*}

\paragraph{Benchmarks} 
We choose five benchmarks for experiments, where HelloBench includes two subsets, as shown in Table~\ref{tab: benchmark}. See details in Appendix \ref{sec:bench_detail}.
% Ruler是我们唯一能获取code并运行的training-based的baseline。

\paragraph{Baselines} 
\begin{itemize}[leftmargin=20pt]
\setlength{\itemsep}{0pt}
\setlength{\parsep}{0pt}
\setlength{\parskip}{0pt}
\vspace{-5pt}
\item \textbf{Ruler} \citep{ruler}: 
A training-based\footnote{Ruler is the only training-based baseline for which we can find that releases the code and training set.} method that defines length control templates to regulate generation at the range level.
% A training-based\footnote{Ruler is the only training-based baseline for which we can obtain the code and run it.} method that defines length control templates controlling at the range level.
\item \textbf{Implicit} \citep{longwriter}: 
Conduct a plan-and-generate process  without explicit counting. 
To ensure a fair comparison, the model generates multiple responses until token count outperforms \textsc{MarkerGen} and the candidate with the smallest LCTG error is selected.
% 为了比较的公平性，模型会生成多次responses直到generated tokens数量和\textsc{MarkerGen}对齐，然后选择LCTG error最小的候选。
\end{itemize}

\paragraph{Details}  
% \vspace{-5pt} 
We conduct extensive experiments 
using Qwen2.5 series (Qwen2.5-7B/14B/32B-Instruct) \citep{qwen25} and the Llama3.1 series (Llama-3.1-8B/70B-Instruct) \citep{llama31}, with sampling temperature as 0.5.
% 我们在HelloBench的Open-ended QA subset上测试在 coarse-grained length constraints下的性能，在其他benchmarks上测试precise length constraints下的LCTG error rate following Eq.~\eqref{eq:E}。
We experiment under coarse-grained length constraints on the Open-ended QA subset of HelloBench and assess the LCTG error rate under precise length constraints on other benchmarks, following Eq.~\eqref{eq:E}.
To evaluate the text quality, we use GPT-4o mini \citep{gpt4o} as the judge, with a calibration algorithm to mitigate the length bias \citep{llmasjudge} (See details in Appendix~\ref{sec:exp details}).
For precise constraints, we set the length of ground truth response as desired target length.
We run each setting for three times and report the average results.

\subsection{Main Results} \label{sec:performance}
As shown in Table \ref{tab:Overall}, the commonly used two-stage implicit counting baseline results in a substantial LCTG error rate $E$ of 18.32\% on average, even if the best response  is chosen across multiple attempts.
This intuitively demonstrates the impact of the inherent limitations of LLM's LCTG sub-capability.
    The training-based baseline Ruler, as observed in our preliminary experiments (Appendix \ref{sec:Training-based methods performance}), benefits from training on test sets that matches the training domain, while performs poorly on our evaluated benchmarks, highlighting its limited generalizability.
In comparison, under strict length constraints, \textsc{MarkerGen} achieves an absolute reduction of 12.57\% in $E$ relative to the implicit baseline, bringing the final error down to just 5.57\%.
In terms of text quality, by decoupling length modeling and semantic modeling during the generation process and employing the decaying insertion strategy to minimize the damage caused by length constraints to semantic integrity, \textsc{MarkerGen} achieves a higher $S$ in average.
Meanwhile, this performance is achieved with only 64\% of the tokens used by the baseline.

\begin{figure*}[t]
  % \vspace{-1cm} % 负间距让图片靠近顶部
  \includegraphics[width=\textwidth, height=0.25\textheight]{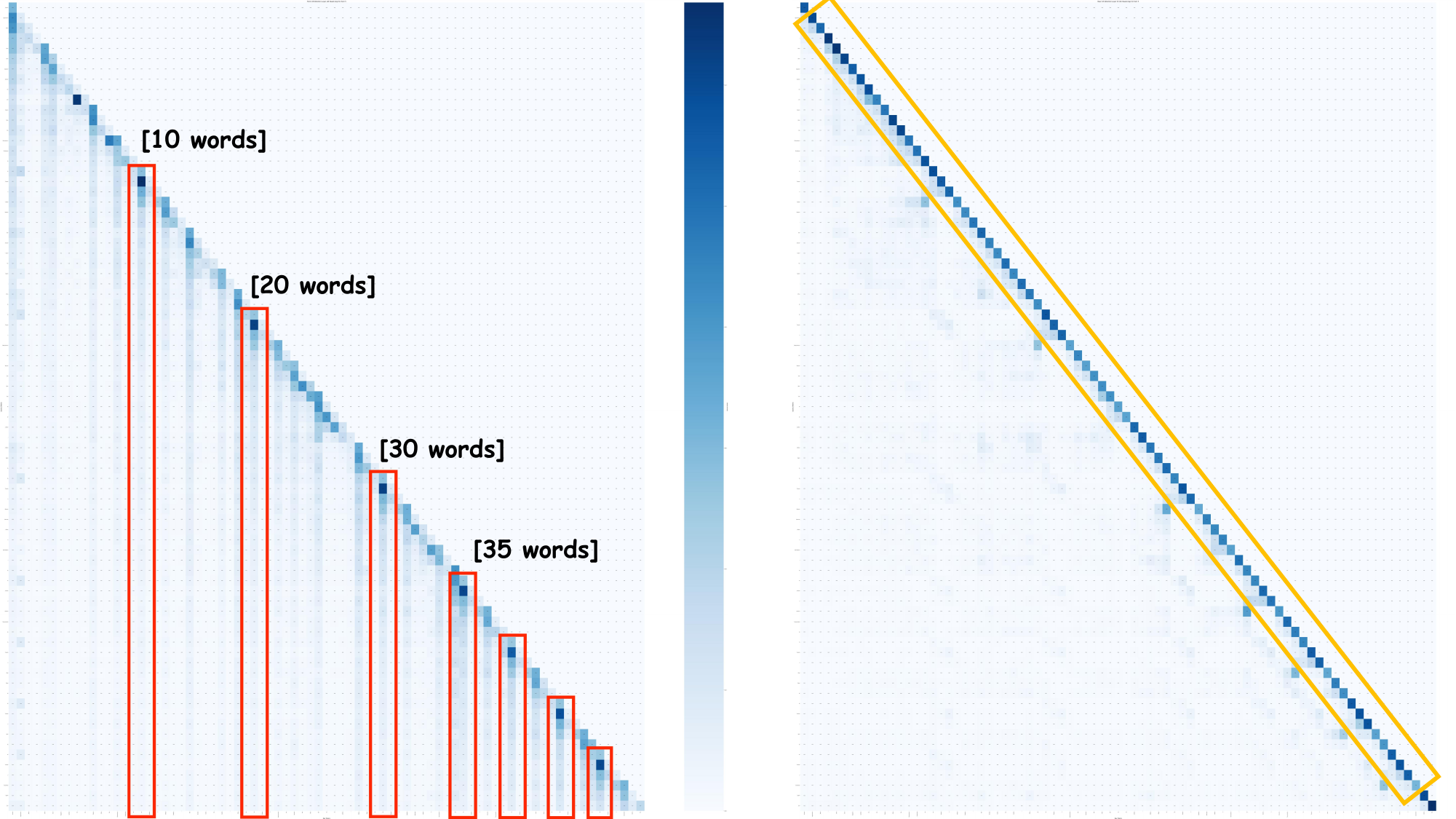}
  \caption{Attention matrices of the first (left) and last (right) layers.}
 % 负间距让图片靠近顶部
  \label{fig:Figure7} 
  \vspace{-4pt}
\end{figure*}

\begin{table*}[t]
  \centering
  \small
  \renewcommand{\arraystretch}{0.9}
   % \normalsize
  \begin{tabular}{l *{6}{cc} }
  % \small
    \toprule
\multirow{3}{*}{Variants} & \multicolumn{12}{c}{Marker Insertion Interval $n$}  \\
    \cmidrule(lr){2-13}
      & \multicolumn{2}{c}{1} & \multicolumn{2}{c}{4} & \multicolumn{2}{c}{16} & \multicolumn{2}{c}{32} & \multicolumn{2}{c}{64} & \multicolumn{2}{c}{Decaying} \\
    \cmidrule(lr){2-13}
      & $E$ ($\downarrow$)& \textcolor{gray}{$S$ ($\uparrow$)} & $E$ ($\downarrow$)& \textcolor{gray}{$S$ ($\uparrow$)}& $E$ & \textcolor{gray}{$S$ ($\uparrow$)} & $E$ ($\downarrow$)& \textcolor{gray}{$S$ ($\uparrow$)} & $E$ ($\downarrow$)& \textcolor{gray}{$S$ ($\uparrow$)} & $E$ ($\downarrow$)& \textcolor{gray}{$S$ ($\uparrow$)} \\
    \midrule
        \textit{w/o Tool}  & 15.53 & \textcolor{gray}{4.28} & 32.50 & \textcolor{gray}{4.29} & 34.64   & \textcolor{gray}{4.46}   & 32.50 & \textcolor{gray}{4.48} & 20.44 & \textcolor{gray}{4.58} & --   & -- \\
      \textit{Two Stage} & 3.10 & \textcolor{gray}{4.03} & 1.49  & \textcolor{gray}{4.03} & 4.04 & \textcolor{gray}{4.20} & 3.26  & \textcolor{gray}{4.23} & 3.93  & \textcolor{gray}{4.32} & 2.66 & \textcolor{gray}{4.28} \\
      \textit{Three Stage}     & 4.84 & \textcolor{gray}{4.28} & 4.20 & \textcolor{gray}{4.29} & 4.89 & \textcolor{gray}{4.45} & 5.45  & \textcolor{gray}{4.48} & 5.18  & \textcolor{gray}{4.57} & 4.48 & \textcolor{gray}{4.54} \\
    \bottomrule
  \end{tabular}%
  \caption{\label{tab:ablation}Ablation studies on key components.}
  \vspace{-10pt}
\end{table*}

\subsection{Generalizability}  
\label{sec:Generalizability}
\paragraph{Across LLMs and Tasks.}
Table~\ref{tab:Overall} demonstrates the strong generalizability of \textsc{MarkerGen} to LLMs and generation tasks.
% Table~\ref{tab:Overall} 展示了在precise length constraints下\textsc{MarkerGen}对于LLMs和generation task的strong generalizability。
% 我们进一步探究其在
\vspace{-5pt}
\paragraph{Across Length Scale.}
% Table~\ref{tab:Overall} 展示了在具有不同长度需求(18-1450)的benchmarks上\textsc{MarkerGen}都能表现良好。我们进一步探查在一个benchmark上，扩展长度需求（从100扩大到400），\textsc{MarkerGen} 的error rate整体呈现下降趋势。这是因为auxiliary marker insertion decoding strategy 能够避免模型隐式建模导致的误差累积。因此target 长度越大，相对误差越小。
Table~\ref{tab:Overall} also shows \textsc{MarkerGen}’s strong performance across benchmarks with varying length scale (18-1450). To further investigate, we analyze progressively increasing the target length from 100 to 400. The results in Table~\ref{tab:openqa} show a declining trend in \textsc{MarkerGen}’s error rate, which can be attributed to the auxiliary marker insertion decoding mechanism that prevents error accumulation from implicit modeling.
\vspace{-5pt}
\paragraph{Across Constraint Types.}
In addition to exact length constraints, users may impose range-based limits. We evaluate $E_r$\footnote{$E_r = 1 - \frac{|\{N_{\mathrm{true}} \mid N_{\mathrm{target}}^{\mathrm{min}} \le N_{\mathrm{true}} \le N_{\mathrm{target}}^{\mathrm{max}}\}|}{N_{\mathrm{total}}}$.}, the proportion of responses violating these constraints. Table~\ref{tab:openqa} shows that \textsc{MarkerGen} maintains an $E_r$ below 3\% in all cases, significantly lower than the baseline.
\vspace{-5pt}
\paragraph{Across Lingual.}
We further validate the effectiveness of \textsc{MarkerGen} in Chinese setting on GAOKAO benchmark, as shown in Table~\ref{tab:chinese}.
% 我们进一步在中文benchmark 上验证了\textsc{MarkerGen}在不同语言上的有效性。
% 除了精准的长度需求，有时用户会给出range-based constraints。
% 我们统计该类粗粒度长度限制下，生成的response不满足区间要求的比率$E_r$. 如表所示，在所有的settings下，\textsc{MarkerGen}都展现了低于3\%的$E_r$，显著低于baseline。
% We define the rejection rate \( E_r \), which is the proportion of items that do not meet the specified range requirements:
% \begin{equation}
% E_r = 1 - \frac{|\{i \in I \mid l_{\text{min}} \leq \text{len}(i) \leq l_{\text{max}}\}|}{|I|}
% \label{eq:Er}
% \end{equation}

% We evaluate the generalizability of \textsc{MarkerGen} across 4 tasks with varying length scales ranging from 18 to 1450 words for fine-grained length-controlled settings, but also in more practical and widely applicable scenarios, such as high-frequency length constraints (e.g., "Please answer in 200/300 words") and range-based length constraint settings (e.g., "Answer length should be between 100-150 words"). In addition, we test \textsc{MarkerGen} on Chinese language generation tasks to further assess its practical utility and generalization ability. For range-based constraints, we define the rejection rate \( E_r \), which is the proportion of items that do not meet the specified range requirements:

% \begin{equation}
% E_r = 1 - \frac{|\{i \in I \mid l_{\text{min}} \leq \text{len}(i) \leq l_{\text{max}}\}|}{|I|}
% \label{eq:Er}
% \end{equation}

% As shown in Tables \ref{tab:openqa} and \ref{tab:chinese}, our method performs well across all tasks, effectively demonstrating its generalizability and practicality.

\subsection{Ablation Studies} \label{sec:ablation}
In this section, we validate the effectiveness of each module in \textsc{MarkerGen} with Qwen2.5-32B-Instruct on TruthfulQA, as shown in Table \ref{tab:ablation}.
% , we replace each step of our default \textsc{MarkerGen} settings with specific variants to measure their individual impact.
\paragraph{Tool Invocation.} 
% 当模型不依靠external tokenizer and counter而被要求自行插入marker时，由于fundamental能力的不足，会导致巨大的（超过15%）error rate。
When the model is required to insert markers independently without relying on an external tokenizer and counter, its fundamental limitations lead to a significant increase in the error rate, exceeding 15\%.
\paragraph{Decaying Interval Marker Insertion.} 
% 在使用固定的marker insertion interval时，由于length modeling和$n$成正比而semantic modeling与$n$成反比（which会引发alignment error），我们观察到LCTG error非常不稳定。相比之下，Decaying Interval Marker Insertion strategy 通过先疏后密的插入策略，在保证了长度显示建模的同时，最大程度上保证了语义完整性，导向了更低的$E$和良好的$S$。
When using a fixed marker insertion interval $n$, since length control is inversely proportional to $n$, while semantic modeling is directly proportional to $n$ (which induces alignment errors), we observe unstable LCTG error rate. In contrast, by adopting a sparse-to-dense insertion approach, the Decaying Interval Marker Insertion strategy ensures explicit length modeling while maximizing semantic integrity, leading to lower $E$ and superior $S$.
\vspace{-5pt}
\paragraph{Three-Stage Decoupled Generation.}
% 两阶段的scheme通过引入插入的length marker相比于implicit baseline能带来更低的LCTG error(8.7-2.66)，但也由于更偏重长度建模而导致了text quality的下降(4.45-4.28)。相比之下，三阶段的scheme通过结耦较好地平衡了semantic和length modeling，带来了both 长度控制和文本质量的提升。
The introduction of explicit length markers in the two-stage scheme leads to a substantial reduction in LCTG error relative to the implicit baseline $(8.7 \rightarrow 2.66)$. However, this scheme places greater emphasis on length modeling, which consequently diminishes text quality $(4.45 \rightarrow 4.28)$. In comparison, the three-stage scheme achieves a better balance by decoupling semantic and length modeling, thereby improving both length control and text quality.

\subsection{Working Mechanism of \textsc{MarkerGen}} \label{sec:analysis}
% \paragraph{Working Mechanism of \textsc{MarkerGen}}
To better understand how LLMs leverage the inserted length markers in \textsc{MarkerGen}, we visualize the attention matrices of the first and last layers of Llama-3.1-8b-Instruct (Figure \ref{fig:Figure7}). In the shallow layers, the attention distribution reveals a clear focus on the length information represented by the length markers (in the red box). As the model progresses to the deeper layers, attention shifts from the length information to the adjacent semantic content (in the orange box). 
% This pattern 说明模型在浅层通过marker进行长度建模，并形成准确的长度信息。而在深层，模型在长度信息的指导下进行语义建模，并最终输出符合长度需求和保障语义完整性的token。
This pattern demonstrates that at shallow layers, the model uses markers to establish length modeling and encode precise length information. At deeper layers, it relies on this length information for semantic modeling, producing tokens that align with the length constraints while maintaining semantic integrity. 
% See more details in Appendix \ref{sec:exp details}.

% This pattern demonstrates the effectiveness of the Marker Insertion Strategy in precise length modeling. Additionally, cross-layer attention analysis reveals a consistent trend: shallow layers prioritize length modeling, while deeper layers focus on semantic generation, specifically targeting relevant output (detailed in Appendix \ref{sec:exp details}). This finding offers valuable insights for future research.

%% file: conclusion.tex
\section*{Conclusions}
To improve the performance of LLMs in length-controllable text generation, we conduct a bottom-up error analysis of relevant sub-abilities. The results reveal that deficiencies in identifying, counting, and aligning are key limitations. To fill this gap, we propose \textsc{MarkerGen}, which leverages external tools to compensate for fundamental deficiencies. Additionally, it introduces Decaying Interval Marker Insertion Strategy to facilitate explicit length modeling and employs Three-Stage Decoupled Generation mechanism to balance semantic coherence and length control. Comprehensive experiments demonstrate the strong generalizability and effectiveness of \textsc{MarkerGen} in enhancing length control and preserving semantic integrity.
% 针对当前LLM在length-controllable text generation任务上表现不佳的问题，我们自底向上地对LLM的相关子能力进行了误差分析。结果表明Identifying，Counting，Aligning 能力的欠缺是主要因素。为此，我们提出\textsc{MarkerGen}方法。其通过外部工具引入解决了fundamental能力的不足，introduce Decaying Interval Marker Insertion Strategy 帮助模型进行显式长度建模，通过Three-Stage Decoupled Generation平衡了语义和长度建模。Comprehensive的实验验证了\textsc{MarkerGen}在增强长度控制和语义完整性上的有效性，以及强大的泛化性。
\section*{Limitations} 
% 我们聚焦于LCTG场景进行了自底向上的子能力分析，并提出\textsc{MarkerGen}方法实现了良好的LCTG效果。
% 我们认为我们的主要Limitation在于：\textsc{MarkerGen}目前只能用于开源模型中。对于闭源的模型，目前还无法应用。我们将公开我们的代码，从而方便愿意适配\textsc{MarkerGen}来提升LCTG效果的闭源模型厂商受益于我们提出的方法。
In this work, we conduct a bottom-up sub-capability analysis in the LCTG ability and propose the \textsc{MarkerGen} method, achieving strong LCTG performance.
One major limitation of \textsc{MarkerGen} is that it is currently only applicable to open-source models and cannot yet be used with closed-source models. To address this, we will release our code, allowing closed-source model providers interested in adapting \textsc{MarkerGen} to benefit from our method in enhancing LCTG performance.

\section*{Ethics Statement}
All of the datasets used in this study were publicly available, and no annotators were employed for our data collection. We confirm that the datasets we used did not contain any harmful content and was consistent with their intended use (research). We have cited the datasets and relevant works used in this study.

\subsection*{Acknowledgments}
This work is supported by Beijing Natural Science Foundation (No.4222037, L181010).

%% file: appendix.tex
\clearpage
\section{Related Work}
\label{sec:related work}

\paragraph{LCTG Methods}
Text length is a fundamental aspect of natural language that carries semantic information, making LCTG a task of balancing length and semantic constraints. Achieving precise length control remains a challenge for LLMs due to limitations in their architecture, such as position encoding \citep{butcher2024precise,kazemnejad2024impact,chang2024language} and decoding mechanisms\citep{huang2025decoding}. Consequently, existing methods focus on injecting length information to help LLMs model length accurately, which can be categorized into training-based and inference-based approaches.

Training-based methods inject varying levels of length signals during fine-tuning or reinforcement learning. For instance, \citet{promptRein, ruler} use prompt templates to teach LLMs the mapping between length and textual content, while \citet{song2024hansel, wang2024positionid} design fine-grained datasets to guide correct length modeling. Other methods, like \citet{lift, promptRein}, utilize reward functions to align length preferences during training. While effective in certain scenarios, these methods suffer from limited generalization across diverse LCTG tasks, including varying length constraints and instructions.Inference-based methods adjust inputs multiple times during generation to inject, such as through prompt-based Automated Revisions and Sample Filtering \citep{retkowski2024zero,juseon2024instructcmp}, or length-controlled importance sampling during decoding \citep{gu2024length}. Although these approaches can better generalize length alignment, they still struggle with achieving precise control.

While both approaches enhance LCTG, they often apply a top-down strategy that lacks deep understanding and targeted enhancement of LCTG sub-capabilities. This limits progress in meeting length constraints accurately. Furthermore, many methods neglect semantic constraints, and injecting length information may degrade text quality. Therefore, we propose \textsc{MarkerGen} to bridge this gap for precise length control and preserving semantic integrity.

\section{Detailed Sub-ability Error analyses in LCTG}
\label{sec:Sub-ability Decomposition}

\begin{figure*}[ht]
\centering
  % \vspace{-1cm} % 负间距让图片靠近顶部
  \includegraphics[width=0.85\textwidth, height=0.3\textheight]{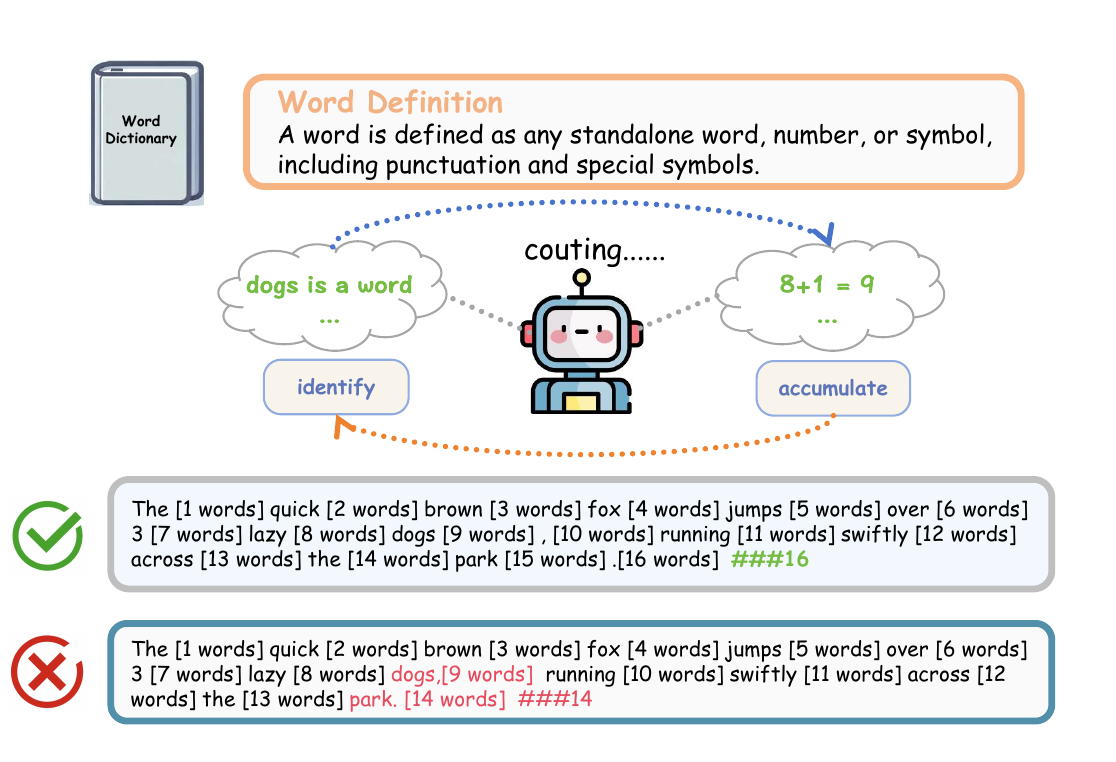}
  \caption{Schematic diagram of counting experiment under the condition of \( n=1 \)}
  \label{fig:Figure8}
\end{figure*}

\subsection{Identifying Error}  
Identifying error refers to the misidentification of fundamental length units. To systematically analyze this error, we design a counting experiment in which the model is prompted to sequentially recognize and accumulate length units, then compare its predicted count with the ground truth. Experimental results confirm that in the one-by-one accumulation setting, counting errors do not occur, meaning that the final length error entirely arises from identifying error (as shown in Figure \ref{fig:Figure8}).

\subsection{Counting Error}  
\begin{figure*}[t]
    % 下面两张图并排
    \begin{minipage}{0.51\textwidth}
        \centering
        \includegraphics[height=5.3cm]{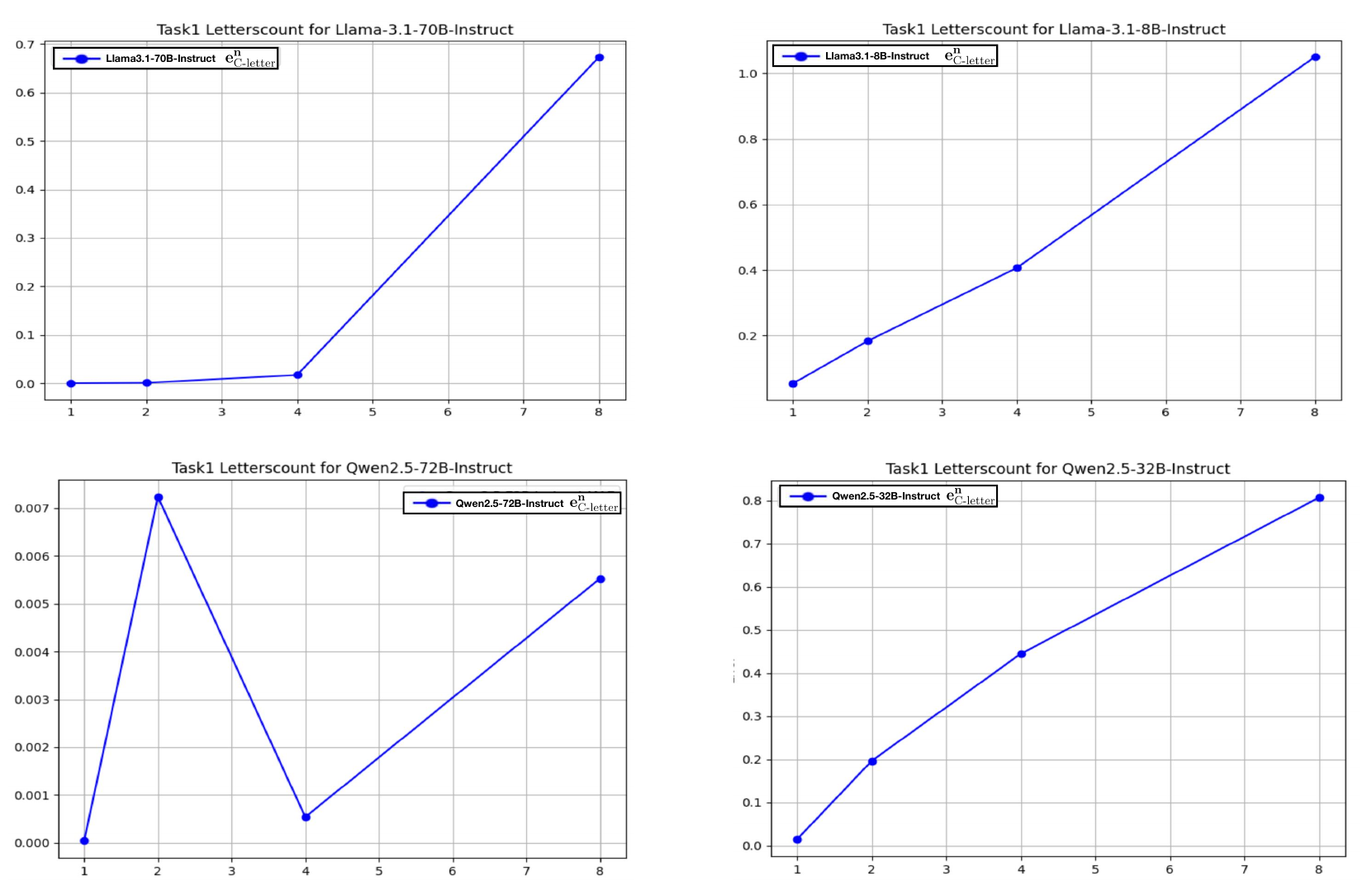}
    \end{minipage}
    \hfill
    \begin{minipage}{0.51\textwidth}
        \centering
        \includegraphics[height=5.3cm]{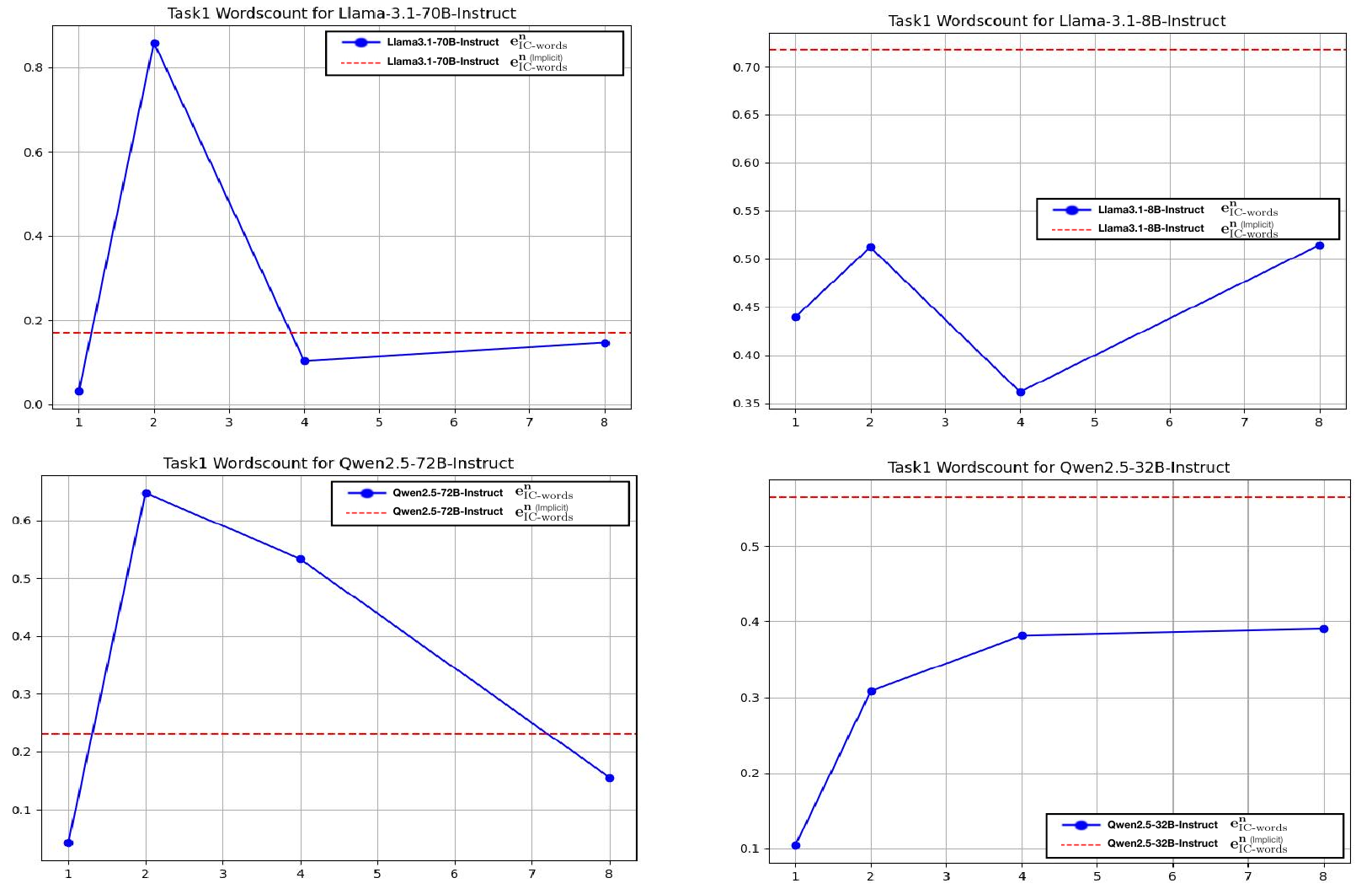}
    \end{minipage}

    \caption{Error analyses of fundamental abilities in LCTG on CNN/DailyMail.}
    \label{fig:Figure9}
\end{figure*}

Counting error refers to the inaccurate enumeration of units in a given sequence, leading to deviations from the intended length. Therefore, in the setting where \(n > 1\) in the counting experiment, the final counting result error is caused by both identifying error and counting error. In this case, counting error can be decoupled by resolving identifying errors in the accumulation process, where errors result from the accumulation step.We also conducted the same counting experiment as in Section \ref{sec:counting_error} on the CNN/DailyMail summarization dataset, as shown in Figures \ref{fig:Figure9}.

From the figure, we can further validate the same conclusions as in Findings \ref{find1}, \ref{find2}, \ref{find3}, and \ref{find4} in Section \ref{sec:Decomposing Length Control Errors}, revealing that the length errors in the generated results of the LCTG task stem from significant errors in the LLM's perception and modeling of length.

\subsection{Planning Error}  
\label{sec:appen_planning}
Planning error refers to the misallocation of word counts across different sections, leading to a discrepancy from target length.The planning ability of LLMs encompasses not only length planning but also semantic planning. To effectively assess the quality of LLM’s semantic planning, we use Qwen-Plus \citep{qwen25} as the judge, with a scoring range of [1, 5]. The specific evaluation prompt is as follows:

\begin{quote}
You are tasked with evaluating the quality of a generated answer plan for a TruthfulQA question. The evaluation should focus on the truthfulness, logical coherence, and adherence to the given prompt and instructions. Rate the answer plan on a 5-point scale as follows:

\begin{itemize}
    \item \textbf{5: Outstanding} - The plan is highly truthful, logically coherent, and perfectly adheres to the prompt and instructions.
    \item \textbf{4: Very Good} - The plan is mostly truthful and coherent, with only minor issues in details or adherence to instructions.
    \item \textbf{3: Good} - The plan is acceptable but has noticeable shortcomings in truthfulness or coherence.
    \item \textbf{2: Poor} - The plan has significant issues in truthfulness or logical coherence and does not adequately follow the instructions.
    \item \textbf{1: Unacceptable} - The plan is largely untruthful, incoherent, or fails to follow the prompt instructions entirely.
\end{itemize}

Please provide the overall score in the following format: \texttt{\#\#\#score X}

\textbf{Question:}

\texttt{+ prompt}

\textbf{Generated Answer Plan:}

\texttt{+ generated\_plan}

Evaluate the answer plan based on the above criteria.
\end{quote}

Since the LCTG task requires meeting both length and semantic constraints, utilizing the LLM's superior planning ability for explicit planning before generation, as opposed to direct generation, helps to clearly define the modeling space for length and the semantic extension range. This not only contributes to improved text generation quality but also reduces length errors.

\begin{figure}[h]
  \includegraphics[width=\columnwidth,height=4.5cm]{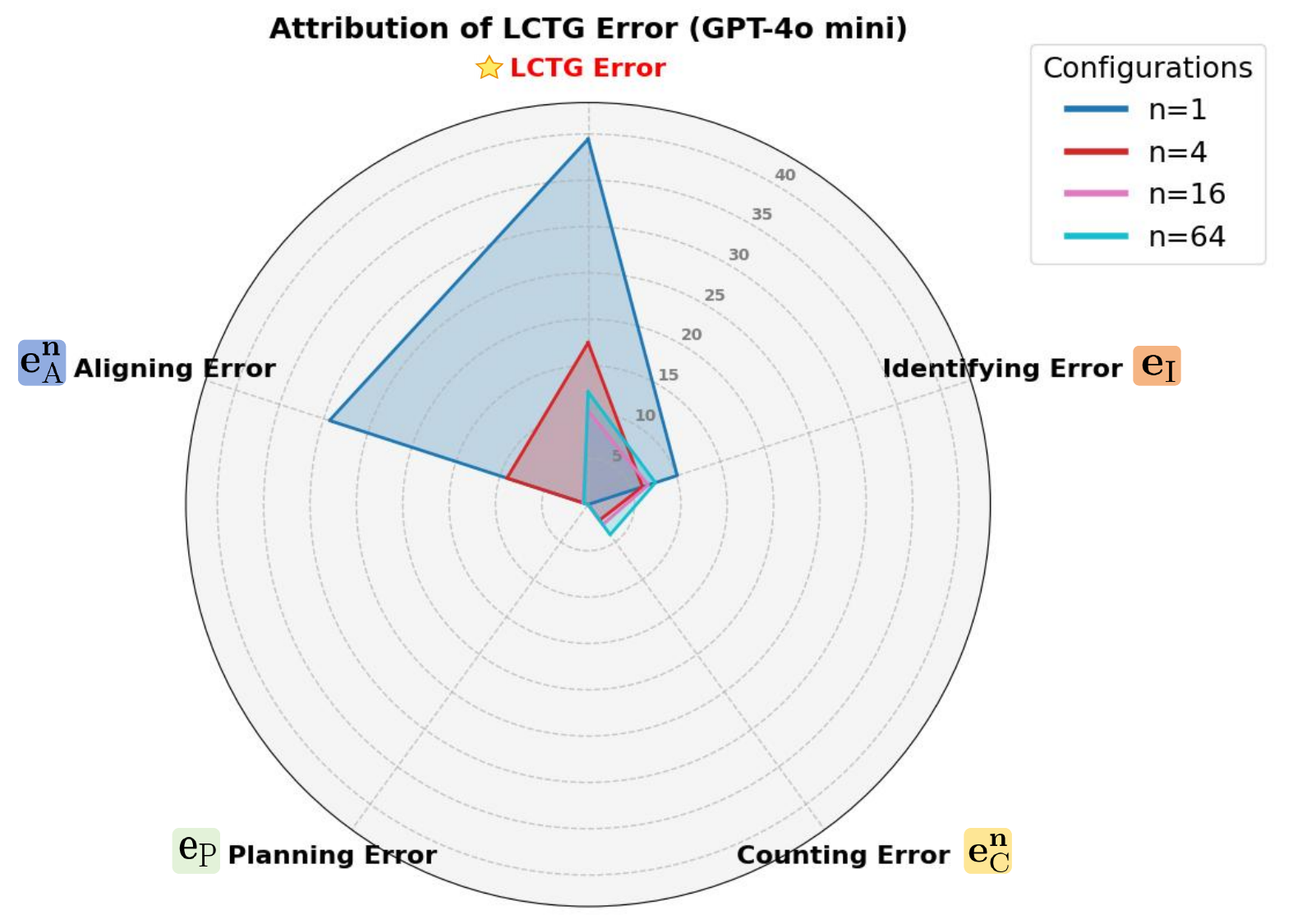}
  \caption{Absolute contribution of LCTG sub-capability deficiency of GPT 4o-mini.}
  \label{fig:Figure10}
\end{figure}

\subsection{Aligning Error}

Aligning error refers to the discrepancy between the model’s perceived length and the target length, arising from the challenge of maintaining semantic integrity while adhering to length constraints.As shown in Figure \ref{fig:Figure4}, aside from Finding \ref{find7}, we observe significant differences in aligning error across models. Qwen2.5-32B-Instruct and GPT-4o mini exhibit larger alignment errors under fine-grained length modeling. As discussed in Section \ref{sec:Decomposing Length Control Errors}, “length estimation acts as a real-time constraint, dynamically regulating further extension. Ultimately, the model strives to align the length constraints while preserving semantic integrity.” High-frequency length perception updates pose greater challenges for the natural expansion of the semantic space, which explains why some models with weaker robustness in semantic expansion show significant alignment errors. These errors become a primary source of LCTG inaccuracies (as shown in Figure \ref{fig:Figure10}). This further emphasizes that LCTG is a task of balancing length and semantic constraints.

\subsection{LCTG Error}

Based on the above decomposition of sub-abilities in LCTG and the corresponding error analysis, we can clearly quantify the contribution of each decoupled error to the final LCTG error. As shown in Figure \ref{fig:Figure0}, the quantification results in the right figure represent the average values of four models under various \ n \
conditions. The conclusion we can draw is that the primary cause of significant length errors in current mainstream LLMs on LCTG tasks is the lack of bottom-up identification and counting capabilities required for accurate length modeling.

\section{Exploration of Interval Marker Insertion Strategy Variants}

\subsection{Length Marker Forms}
\label{sec:LM_form}

We explored the impact of different forms of length marker insertion on performance, such as the number of words generated "[k]", the semantic marker "[k words]", and the remaining words to be generated "[ $N_{\mathrm{target}}$ - k]" (remaining words). As shown in Table \ref{tab:marker_comparison}, we used Llama-3.1-8B-Instruct on the CNN/DailyMail dataset to investigate the effects of various marker forms under multiple \(n\) conditions on generation error and text quality. The results show that using a semantic length marker representing the number of words generated achieved the best performance in both length error and text quality.

\begin{table}[ht]
\centering
   \renewcommand\arraystretch{1.0}
  \setlength{\tabcolsep}{1.5em} 
\begin{tabular}{l c c}
\toprule
\textbf{Marker Form} & $E$ ($\downarrow$)  & $S$ ($\uparrow$) \\
\midrule
$[k]$ & 18.28 & 3.10 \\
$[k\ \text{words}]$ & \textbf{15.74} & \textbf{3.14} \\
$[N_{\mathrm{target}} - k]$ & 27.92 & 3.09 \\
\bottomrule
\end{tabular}
\caption{Comparison of Length Marker Forms and Their Performance}
\label{tab:marker_comparison}
\end{table}

\section{Detailed Benchmarks Introduction}
\vspace{-4pt}
\label{sec:bench_detail}
The benchmarks used in our experiments are as follows:
\vspace{-5pt}  
\begin{itemize}
    \setlength{\itemsep}{-2pt} % 调
    \item \textbf{CNN/DailyMail}\citep{CNN/DailyMail}: A summarization dataset of news articles, with 500 randomly sampled items. (\textit{18-165 words})
    \item \textbf{HANNA}\citep{HANNA}: A long-form story generation dataset with 200 selected items. (\textit{139-995 words})
    \item \textbf{TruthfulQA}\citep{TruthfulQA}: A benchmark for factual accuracy in open-domain QA. (\textit{101-294 words})
    \item \textbf{HelloBench}\citep{HelloBench}: A long-text generation benchmark. We selected subsets from \textit{heuristic text generation} (e.g., argumentative and roleplaying writing, covering five types) and \textit{open-ended QA} (spanning ten domains). (\textit{489-1450 words})
    \item \textbf{GAOKAO-Bench}\citep{gaokao}: A benchmark collected from the Chinese college entrance examination (GAOKAO). We selected the \textit{2010-2022 History Open-ended Questions} subset. (\textit{71-901 words})
\end{itemize}

\section{Detailed Experimental Results}
\label{sec:exp details}

\begin{table*}[ht]
  \centering
  \small
  \setlength{\tabcolsep}{1.30em} 
\renewcommand{\arraystretch}{1.00}
\begin{tabular}{ccccc}
\toprule
\multicolumn{5}{c}{\textbf{TLG dataset}} \\
\midrule
benchmark & Method & Models & $PM$ ($\uparrow$) & $FM$ ($\uparrow$)\\
\midrule
\multirow{2}{*}{TLG dataset} 
& before training & Llama-3.1-8B-Instruct & 5.55 & 10.20  \\
& RULER &Llama-3.1-8B-Instruct-ruler & 41.75 & 55.10  \\
\midrule
\multicolumn{5}{c}{\textbf{Precise Length Constraint Benchmarks }} \\
\midrule
Benchmarks & Methods & Models &  $E$ ($\downarrow$)  & $S$ ($\uparrow$)\\
\midrule
\multirow{2}{*}{CNN/DailyMail}  
& Ruler & Llama-3.1-8B-Instruct-ruler & 78.21 & 3.10  \\
& \textsc{MarkerGen} & Llama-3.1-8B-Instruct & 3.36 & 3.18 \\
\midrule
\multirow{2}{*}{HANNA}  
& Ruler & Llama-3.1-8B-Instruct-ruler & 68.21 & 2.87 \\
& \textsc{MarkerGen} & Llama-3.1-8B-Instruct & 2.98 & 3.60 \\
\midrule
\multirow{2}{*}{TruthfulQA}  
& Ruler & Llama-3.1-8B-Instruct-ruler & 44.93 & 3.27 \\
& \textsc{MarkerGen} & Llama-3.1-8B-Instruct & 3.82 & 4.25 \\
\midrule
\multirow{2}{*}{Heuristic Generation}   
& Ruler & Llama-3.1-8B-Instruct-ruler & 66.17 & 2.94 \\
& \textsc{MarkerGen} & Llama-3.1-8B-Instruct & 6.03 & 4.03 \\
\bottomrule
\end{tabular}
\caption{\label{tab:ruler_results}
Evaluation of the training-based method \textsc{Ruler} on its own test set and four generalization benchmarks.}

\end{table*}

\begin{table}[ht]
\centering
    \small
   \renewcommand\arraystretch{1.2}
  \setlength{\tabcolsep}{0.6em} 
\begin{tabular}{l l c c}
\toprule
\textbf{Methods} & \textbf{Models} & $E$ ($\downarrow$) & $S$ ($\uparrow$)\\
\midrule
Implicit & Qwen2.5-14B-Instruct& 27.41 & 3.47 \\
\textsc{MarkerGen} & Qwen2.5-14B-Instruct & 7.71 &3.55  \\
\bottomrule
\end{tabular}
\caption{GAOKAO-History Chinese Dataset Results}
\label{tab:chinese}
\end{table}

\subsection{Performance and Generalization Study of Training-based Methods}
\label{sec:Training-based methods performance}
To investigate the performance and generalization of training-based methods in diverse, real-world LCTG task scenarios, we selected Ruler, a training-based method that defines length control templates to regulate generation at the range level. This choice is based on the fact that Ruler is the only training-based baseline for which the code and training set are publicly available. We followed the exact setup provided in the repository and verified the correctness of our replication by achieving significant performance improvements on the given test set, as shown in Table \ref{tab:ruler_results}.

The test set of Ruler adopts two custom evaluation metrics: Precise Match (PM) and Flexible Match (FM). PM requires the output length to fall within a narrow target interval, while FM allows a broader tolerance range. The metrics are defined as:
\[
PM = \frac{1}{N} \sum_{i=1}^{N} 1 \left( \text{lb}^{P}_{\text{TL}_i} < L(c_i) \leq \text{ub}^{P}_{\text{TL}_i} \right)
\]
\[
FM = \frac{1}{N} \sum_{i=1}^{N} 1 \left( \text{lb}^{F}_{\text{TL}_i} < L(c_i) \leq \text{ub}^{F}_{\text{TL}_i} \right)
\]
where $L(c_i)$ is the length of the $i$-th generated output, and $\text{TL}_i$ denotes the corresponding target length.

Next, we tested the trained model, referred to as Llama-3.1-8B-Instruct-ruler, across four selected benchmarks with varying tasks, length scales, and instructions, under cost-alignment conditions. The experimental results revealed substantial errors and a decline in text quality, even when compared to the implicit method's results without training (as shown in Table \ref{tab:Overall}). This finding demonstrates the limited generalization capability of the method, highlighting its struggle to cope with the complexity and diversity of real-world LCTG scenarios.

\subsection{Length Bias Correction in LLMs-as-a-Judge}

It has been demonstrated that LLMs-as-a-judge exhibit a noticeable length bias \citep{li2024generation, gu2024survey}. To evaluate the quality of generated text objectively and accurately for LCTG tasks, it is essential to correct for this length bias. We adopt the length-controlled AlpacaEval \citep{dubois2024length} and \citet{yuan2025llm}.

To derive unbiased judge scores, we use a Multiple Regression model. Specifically, we set the judge score as the dependent variable, with the generator categories as dummy variables, and the length of the generated text as a covariate. The model is formulated as follows:

\begin{equation}
\small
f(i) = \beta_0 + \beta_{M} \cdot C(\text{Method}) + \beta_{\text{m}} \cdot C(\text{Model}) + \beta_{\text{l}} \cdot \text{Length} + \epsilon
\label{equation: bias}
\end{equation}

Where \( f(i) \) denotes the judge score for the generated text \( \mathcal{G}_{i} \), \( C(\text{Method}) \) and \( C(\text{model}) \) are categorical variables representing the method and the model used, respectively, and \( \text{Length} \) is the actual length of the generated text. The coefficients \( \beta_{\text{M}} \), \( \beta_{\text{m}} \), and \( \beta_{\text{l}} \) are used to adjust the raw judge score \( f(i) \), effectively removing length bias by setting it to zero. These adjusted scores, free from length bias, serve as the metrics for faithfulness and alignment.

\begin{figure*}[t]
\centering
  % \vspace{-1cm} % 负间距让图片靠近顶部
  \includegraphics[width=0.9\textwidth, height=0.3\textheight]{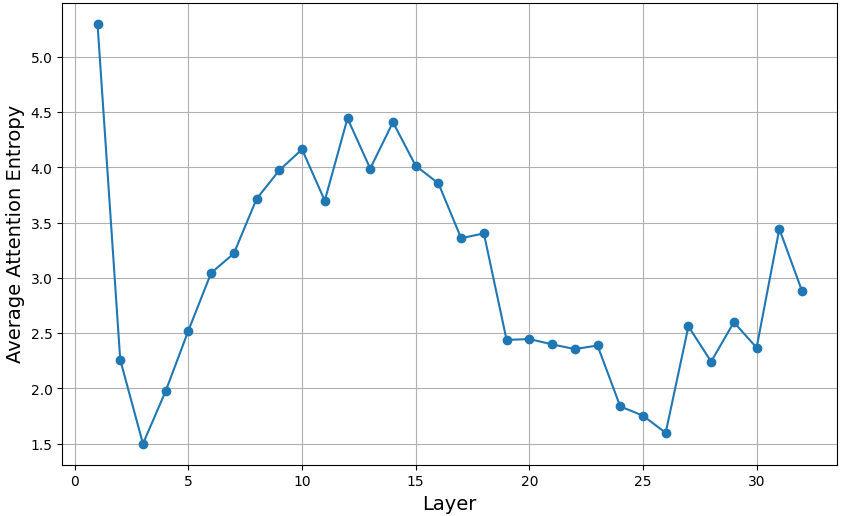}
  \caption{Attention Entropy across layers.}
  \vspace{-0.5cm} % 负间距让图片靠近顶部
  \label{fig:Figure11}
\end{figure*}

\subsection{Win-rate and LLM-as-Judge Based Evaluation}
\label{sec:winrate-eval}

To provide a more robust and trustworthy evaluation of generation quality, we complement our main evaluation protocol with a win-rate-based assessment and multi-LLM judgment analysis. Specifically, we adopt the pairwise evaluation framework from AlpacaEval 2.0~\cite{dubois2024length}, using GPT-4o-mini as the automatic judge. The generated outputs from the Implicit baseline and our proposed \textsc{MarkerGen} method are compared against reference model Gemini-2.0 Flash\cite{google2024geminiflash} across four benchmarks. As shown in Table~\ref{tab:winrate_results}, \textsc{MarkerGen} consistently achieves higher win rates across most configurations, verifying the effectiveness of the Three-Stage Decoupled Generation in improving response quality.

To further confirm the reliability of using LLMs as judges, we conducted a human agreement study. We randomly sampled 200 pairwise comparisons and observed a strong Spearman's correlation coefficient of 0.8794 between GPT-4o-mini judgments and human annotations, indicating high consistency.

In addition, we report point-wise quality scores evaluated by three different LLMs: GPT-4o, Gemini-2.0-Flash (\texttt{thinking-exp}), and GPT-4o-mini. Tables~\ref{tab:cnn_results},~\ref{tab:hanna_results},~\ref{tab:truthfulqa_results}, and~\ref{tab:heuristic_results} summarize these results across the CNN/DailyMail, HANNA, TruthfulQA, and Heuristic Generation benchmarks, respectively. Each table reports the average score ($S_{\text{avg}}$), variance ($\sigma^2$), and model sizes. \textsc{MarkerGen} consistently improves average scores across all model sizes and datasets, while maintaining comparable or lower variance, further reinforcing its robustness and generalizability.

\begin{table*}[ht]
\centering
\small
\setlength{\tabcolsep}{0.6em}
\renewcommand{\arraystretch}{1.1}
\begin{tabular}{cc|c|cccc}
\toprule
\textbf{Model Series} & \textbf{Size} & \textbf{Method} & \textbf{CNN/DailyMail} & \textbf{HANNA} & \textbf{TruthfulQA} & \textbf{Heuristic Gen.} \\
\midrule
\multirow{6}{*}{\textbf{Qwen2.5}} 
& 32B & Implicit    & 15.80\% & 6.50\%  & 28.25\% & 1.63\%  \\
&     & \textsc{MarkerGen} & \textbf{16.43\%} & \textbf{23.00\%} & \textbf{36.91\%} & \textbf{5.69\%} \\
& 14B & Implicit    & 9.40\%  & 7.04\%  & \textbf{28.87\%} & 1.63\%  \\
&     & \textsc{MarkerGen} & \textbf{11.60\%} & \textbf{11.00\%} & 24.12\% & \textbf{3.25\%} \\
& 7B  & Implicit    & 4.20\%  & 1.51\%  & 23.24\% & 0.00\%  \\
&     & \textsc{MarkerGen} & \textbf{14.80\%} & \textbf{16.67\%} & \textbf{25.62\%} & \textbf{4.07\%} \\
\midrule
\multirow{4}{*}{\textbf{Llama3.1}} 
& 70B & Implicit    & 7.40\%  & 2.50\%  & \textbf{33.81\%} & 0.00\%  \\
&     & \textsc{MarkerGen} & \textbf{9.02\%} & \textbf{6.03\%} & 30.31\% & \textbf{1.63\%} \\
& 8B  & Implicit    & \textbf{12.27\%} & 3.02\%  & 19.90\% & 0.00\%  \\
&     & \textsc{MarkerGen} & 10.69\% & \textbf{4.15\%} & \textbf{20.52\%} & 0.00\%  \\
\bottomrule
\end{tabular}
\caption{Win rates (\%) of Implicit baseline and \textsc{MarkerGen} compared against Gemini-2.0 Flash using GPT-4o-mini as judge across four benchmarks. Higher win rates indicate better generation quality.}
\label{tab:winrate_results}
\end{table*}

\begin{table*}[ht]
\centering
\small
\setlength{\tabcolsep}{0.7em}
\renewcommand{\arraystretch}{1.1}

\begin{tabular}{cc|c|rrrrr}
\toprule
\textbf{Model Series} & \textbf{Size} & \textbf{Method} & \textbf{S\_4o} & \textbf{S\_gemini} & \textbf{S\_4omini} & \textbf{S\_avg} & \boldmath{$\sigma^2$} \\
\midrule
\multirow{6}{*}{\textbf{Qwen2.5}}
& 32B & Implicit    & 3.02 & 3.64 & 3.21 & 3.29 & 0.067 \\
&     & \textsc{MarkerGen} & \textbf{3.05} & \textbf{3.67} & \textbf{3.25} & \textbf{3.32} & 0.066 \\
& 14B & Implicit    & 2.92 & 3.55 & 3.15 & 3.21 & 0.099 \\
&     & \textsc{MarkerGen} & \textbf{2.93} & \textbf{3.58} & \textbf{3.16} & \textbf{3.22} & 0.092 \\
& 7B  & Implicit    & 2.68 & 3.39 & 3.04 & 3.04 & 0.125 \\
&     & \textsc{MarkerGen} & \textbf{2.83} & \textbf{3.45} & \textbf{3.07} & \textbf{3.12} & 0.092 \\
\midrule
\multirow{4}{*}{\textbf{Llama3.1}}
& 70B & Implicit    & 2.84 & 3.55 & 3.09 & 3.16 & 0.109 \\
&     & \textsc{MarkerGen} & \textbf{3.08} & \textbf{3.69} & \textbf{3.36} & \textbf{3.38} & 0.068 \\
& 8B  & Implicit    & 2.81 & 3.37 & 3.04 & 3.07 & 0.067 \\
&     & \textsc{MarkerGen} & 2.78 & \textbf{3.38} & \textbf{3.18} & \textbf{3.11} & 0.074 \\
\bottomrule
\end{tabular}
\caption{Evaluation scores of \textsc{MarkerGen} and Implicit baseline on CNN/DailyMail benchmark using three judges (GPT-4o, Gemini-2.0 Flash, GPT-4o-mini). }
\label{tab:cnn_results}
\end{table*}

\begin{table*}[ht]
\centering
\small
\setlength{\tabcolsep}{0.7em}
\renewcommand{\arraystretch}{1.1}

\begin{tabular}{cc|c|rrrrr}
\toprule
\textbf{Model Series} & \textbf{Size} & \textbf{Method} & \textbf{S\_4o} & \textbf{S\_gemini} & \textbf{S\_4omini} & \textbf{S\_avg} & \boldmath{$\sigma^2$} \\
\midrule
\multirow{6}{*}{\textbf{Qwen2.5}} 
& 32B & Implicit    & 3.20 & 3.73 & 3.67 & 3.53 & 0.056 \\
&     & \textsc{MarkerGen} & \textbf{3.29} & \textbf{3.75} & \textbf{3.72} & \textbf{3.59} & 0.044 \\
& 14B & Implicit    & 3.15 & 3.61 & 3.55 & 3.44 & 0.041 \\
&     & \textsc{MarkerGen} & \textbf{3.16} & 3.61 & 3.55 & 3.44 & 0.039 \\
& 7B  & Implicit    & 3.09 & 3.31 & 3.47 & 3.29 & 0.024 \\
&     & \textsc{MarkerGen} & \textbf{3.12} & \textbf{3.36} & \textbf{3.50} & \textbf{3.33} & 0.024 \\
\midrule
\multirow{4}{*}{\textbf{Llama3.1}} 
& 70B & Implicit    & 3.18 & 3.66 & 3.61 & 3.48 & 0.047 \\
&     & \textsc{MarkerGen} & \textbf{3.25} & \textbf{3.74} & \textbf{3.63} & \textbf{3.54} & 0.044 \\
& 8B  & Implicit    & 3.13 & 3.45 & 3.54 & 3.37 & 0.030 \\
&     & \textsc{MarkerGen} & \textbf{3.19} & \textbf{3.50} & \textbf{3.60} & \textbf{3.43} & 0.031 \\
\bottomrule
\end{tabular}
\caption{Evaluation scores of \textsc{MarkerGen} and Implicit baseline on HANNA benchmark using three judges (GPT-4o, Gemini-2.0 Flash, GPT-4o-mini). }
\label{tab:hanna_results}
\end{table*}

\begin{table*}[ht]
\centering
\small
\setlength{\tabcolsep}{0.7em}
\renewcommand{\arraystretch}{1.1}

\begin{tabular}{cc|c|rrrrr}
\toprule
\textbf{Model Series} & \textbf{Size} & \textbf{Method} & \textbf{S\_4o} & \textbf{S\_gemini} & \textbf{S\_4omini} & \textbf{S\_avg} & \boldmath{$\sigma^2$} \\
\midrule
\multirow{6}{*}{\textbf{Qwen2.5}} 
& 32B & Implicit    & 4.56 & 3.94 & 4.45 & 4.32 & 0.073 \\
&     & \textsc{MarkerGen} & \textbf{4.63} & \textbf{4.09} & \textbf{4.54} & \textbf{4.42} & 0.055 \\
& 14B & Implicit    & 4.47 & \textbf{3.95} & \textbf{4.44} & \textbf{4.29} & 0.056 \\
&     & \textsc{MarkerGen} & 4.44 & 3.93 & 4.43 & 4.27 & 0.058 \\
& 7B  & Implicit    & 4.24 & 3.72 & 4.29 & 4.09 & 0.066 \\
&     & \textsc{MarkerGen} & \textbf{4.34} & \textbf{3.83} & \textbf{4.33} & \textbf{4.17} & 0.056 \\
\midrule
\multirow{4}{*}{\textbf{Llama3.1}} 
& 70B & Implicit    & 4.53 & 4.09 & 4.46 & 4.36 & 0.038 \\
&     & \textsc{MarkerGen} & \textbf{4.59} & \textbf{4.14} & \textbf{4.48} & \textbf{4.40} & 0.038 \\
& 8B  & Implicit    & \textbf{4.20} & 3.62 & 4.22 & 4.01 & 0.076 \\
&     & \textsc{MarkerGen} & 4.19 & 3.62 & \textbf{4.25} & \textbf{4.02} & 0.080 \\
\bottomrule
\end{tabular}
\caption{Evaluation scores of \textsc{MarkerGen} and Implicit baseline on TruthfulQA benchmark using three judges (GPT-4o, Gemini-2.0 Flash, GPT-4o-mini).}
\label{tab:truthfulqa_results}
\end{table*}

\begin{table*}[ht]
\centering
\small
\setlength{\tabcolsep}{0.7em}
\renewcommand{\arraystretch}{1.1}

\begin{tabular}{cc|c|rrrrr}
\toprule
\textbf{Model Series} & \textbf{Size} & \textbf{Method} & \textbf{S\_4o} & \textbf{S\_gemini} & \textbf{S\_4omini} & \textbf{S\_avg} & \boldmath{$\sigma^2$} \\
\midrule
\multirow{6}{*}{\textbf{Qwen2.5}} 
& 32B & Implicit    & 4.53 & 4.14 & 3.80 & 4.16 & 0.089 \\
&     & \textsc{MarkerGen} & \textbf{4.64} & \textbf{4.22} & \textbf{4.14} & \textbf{4.33} & 0.049 \\
& 14B & Implicit    & 4.42 & 3.98 & 3.80 & 4.07 & 0.068 \\
&     & \textsc{MarkerGen} & \textbf{4.46} & \textbf{4.07} & \textbf{4.00} & \textbf{4.18} & 0.040 \\
& 7B  & Implicit    & 3.97 & 3.70 & 3.42 & 3.69 & 0.050 \\
&     & \textsc{MarkerGen} & \textbf{4.52} & \textbf{4.15} & \textbf{4.13} & \textbf{4.26} & 0.031 \\
\midrule
\multirow{4}{*}{\textbf{Llama3.1}} 
& 70B & Implicit    & 4.47 & 3.98 & 3.74 & 4.07 & 0.093 \\
&     & \textsc{MarkerGen} & \textbf{4.54} & \textbf{4.08} & \textbf{3.98} & \textbf{4.20} & 0.058 \\
& 8B  & Implicit    & 4.21 & 3.89 & 3.72 & 3.94 & 0.041 \\
&     & \textsc{MarkerGen} & \textbf{4.47} & \textbf{4.04} & \textbf{4.03} & \textbf{4.18} & 0.042 \\
\bottomrule
\end{tabular}
\caption{Evaluation scores of \textsc{MarkerGen} and Implicit baseline on Heuristic Generation benchmark using three judges (GPT-4o, Gemini-2.0 Flash, GPT-4o-mini). }
\label{tab:heuristic_results}
\end{table*}

\subsection{Enhanced Ablation Study}

To provide a more comprehensive validation of our method, we extend the original component ablation study (Table~5) by evaluating additional combinations of stages. This enhanced analysis is motivated by the need to isolate and quantify the individual and joint effects of the key stages, as suggested in the review. Table~\ref{tab:extended_ablation} presents results in terms of error score ($E \downarrow$), LLM-evaluated quality score ($S \uparrow$), and cost (relative to baseline, $\delta$).

\begin{table*}[ht]
\centering
\small
\setlength{\tabcolsep}{0.6em}
\renewcommand{\arraystretch}{1.1}
\begin{tabular}{l l c r r r}
\toprule
\textbf{Variant} & \textbf{Components} & \textbf{Interval} & \textbf{$E$ ($\downarrow$)} & \textbf{$S$ ($\uparrow$)} & \textbf{Cost} \\
\midrule
Baseline (0) & Direct Generation & 64 & 30.27 & 4.51 & $\delta$ \\
\midrule
w/o Tool (1) & Plan + Generation & 64 & 27.58 & 4.60 & $1.05\delta$ \\
             &                   & Decaying & 20.44 & 4.58 & $1.56\delta$ \\
\midrule
w Tool (2) & Insertion Strategy & 64 & 5.96 & 4.27 & $0.79\delta$ \\
           &                    & Decaying & 3.58 & 4.27 & $0.69\delta$ \\
\midrule
w/o Tool (w decouple) (3) & Decoupled Generation & 64 & 26.11 & 4.53 & $1.20\delta$ \\
                          &                       & Decaying & 22.87 & 4.52 & $1.11\delta$ \\
\midrule
Two Stage (1+2) & Plan + Insertion Strategy & 64 & 3.93 & 4.32 & $0.84\delta$ \\
                &                           & Decaying & \textbf{2.66} & 4.28 & $0.79\delta$ \\
\midrule
w/o Tool (w plan \& decouple) (1+3) & Plan + Decoupled Generation & 64 & 25.89 & \textbf{4.63} & $1.56\delta$ \\
                                   &                              & Decaying & 23.19 & 4.60 & $1.51\delta$ \\
\midrule
w Tool (w/o plan \& w decouple) (2+3) & Insertion + Decoupled Generation & Decaying & 5.11 & 4.50 & $1.10\delta$ \\
Three Stage (1+2+3) & Plan + Insertion + Decoupled Gen. & Decaying & 4.48 & 4.54 & $1.46\delta$ \\
\bottomrule
\end{tabular}
\caption{Extended ablation study analyzing the impact of different components combinations. $\delta$: cost relative to baseline.}
\label{tab:extended_ablation}
\end{table*}

As shown in Table~\ref{tab:extended_ablation}, the removal of any individual stage leads to a significant increase in error or a drop in quality. Notably, the two-stage variant (1+2) achieves the lowest error (2.66) with a relatively low cost ($0.79\delta$), confirming its effectiveness and efficiency. Meanwhile, the full three-stage setup offers a balanced trade-off, delivering strong performance with moderate cost. These results reinforce the necessity of each stage and validate our full method design.

\subsection{Residual Length Error Analysis in \textsc{MarkerGen}}

This subsection focuses on analyzing the residual length errors in the \textsc{MarkerGen} framework. Building upon the sub-decomposition of LCTG errors presented in Section \ref{sec:Preliminaries}, we eliminate identifying and counting errors through Auxiliary Length Marker Insertion Decoding \ref{sec:Auxiliary Length Marker Insertion Decoding}. Moreover, by employing the Three-Stage Decoupled Generation strategy \ref{sec:Two-Stage High-Quality Text Generation}, we effectively reduce aligning errors, thus improving the robustness of all models in semantic expansion under precise length modeling with explicit length markers. This approach ensures semantic integrity while enhancing text generation quality through a clearer, more in-depth analysis of LLM’s LCTG sub-capabilities. Ultimately, residual LCTG errors are primarily driven by minimal aligning errors.

\subsection{Cross-layer Attention Analysis from the \textsc{MarkerGen} Perspective}

In this section, we perform a cross-layer attention analysis from the \textsc{MarkerGen} perspective. By examining attention patterns across different layers of the model, we aim to gain a better understanding of how length and semantic information are processed at various stages of generation, providing insights into improving the accuracy of LCTG tasks.

\begin{figure*}[t]
\centering
  % \vspace{-1cm} % 负间距让图片靠近顶部
  \includegraphics[width=0.9\textwidth, height=0.3\textheight]{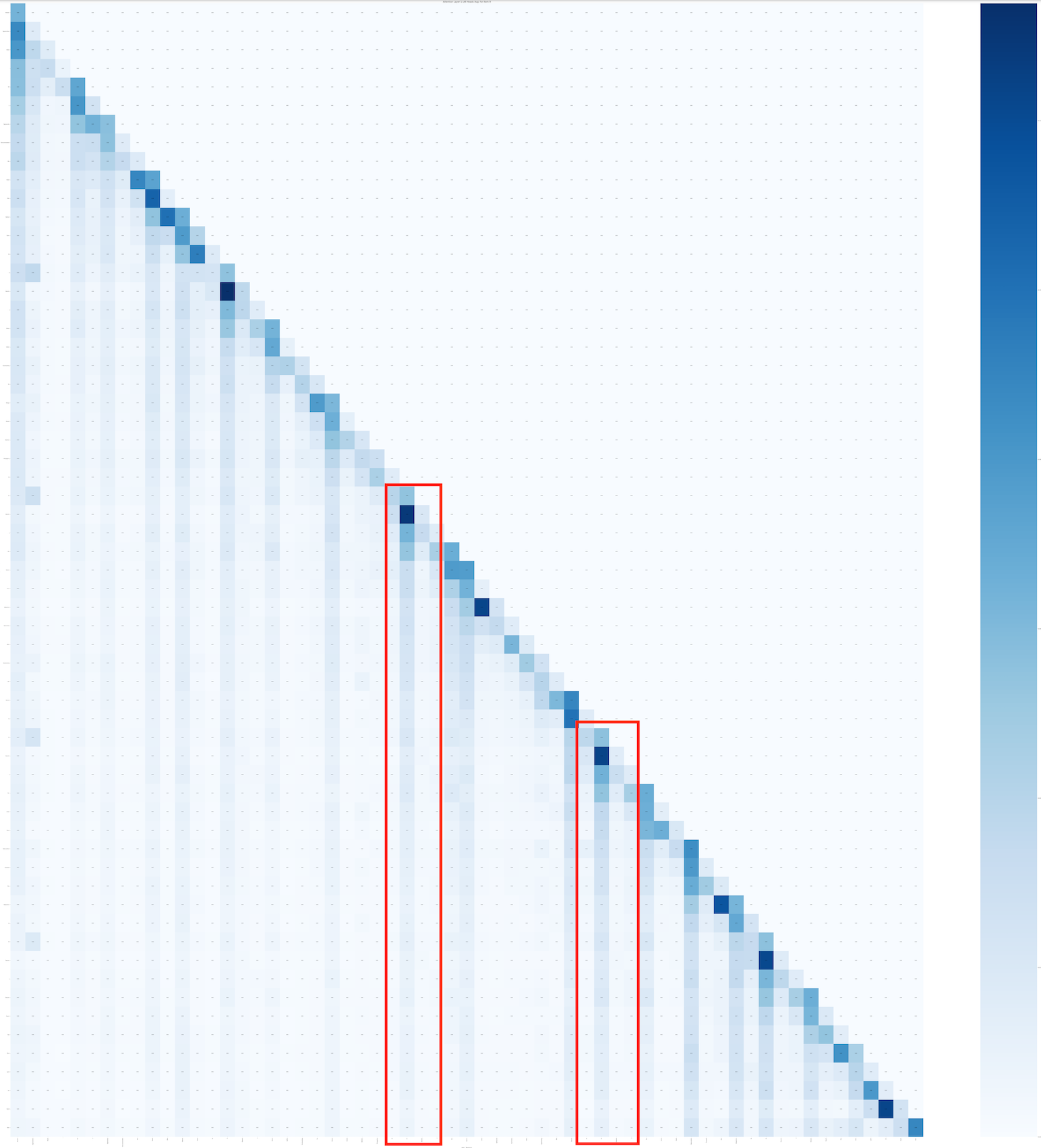}
  \caption{Attention Matrices of the first layers with Insertion Interval \( n=4 \)}
  \vspace{-0.5cm} % 负间距让图片靠近顶部
  \label{fig:Figure12}
\end{figure*}

\begin{figure*}[t]
\centering
  % \vspace{-1cm} % 负间距让图片靠近顶部
  \includegraphics[width=0.9\textwidth, height=0.3\textheight]{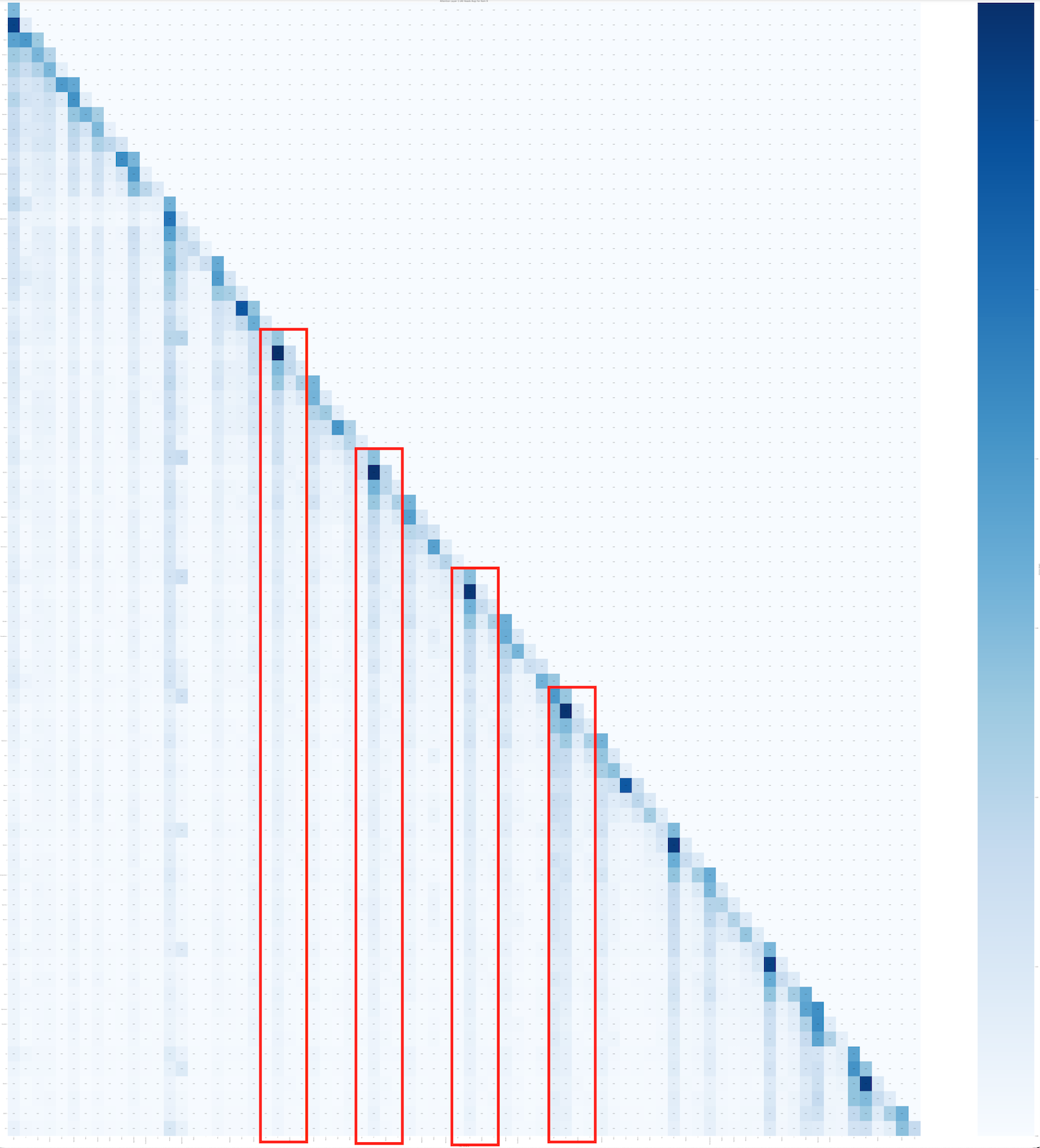}
  \caption{Attention Matrices of the first layers with Insertion Interval \( n=8 \)}
  \vspace{-0.5cm} % 负间距让图片靠近顶部
  \label{fig:Figure13}
\end{figure*}

\begin{figure*}[t]
\centering
  % \vspace{-1cm} % 负间距让图片靠近顶部
  \includegraphics[width=0.9\textwidth, height=0.3\textheight]{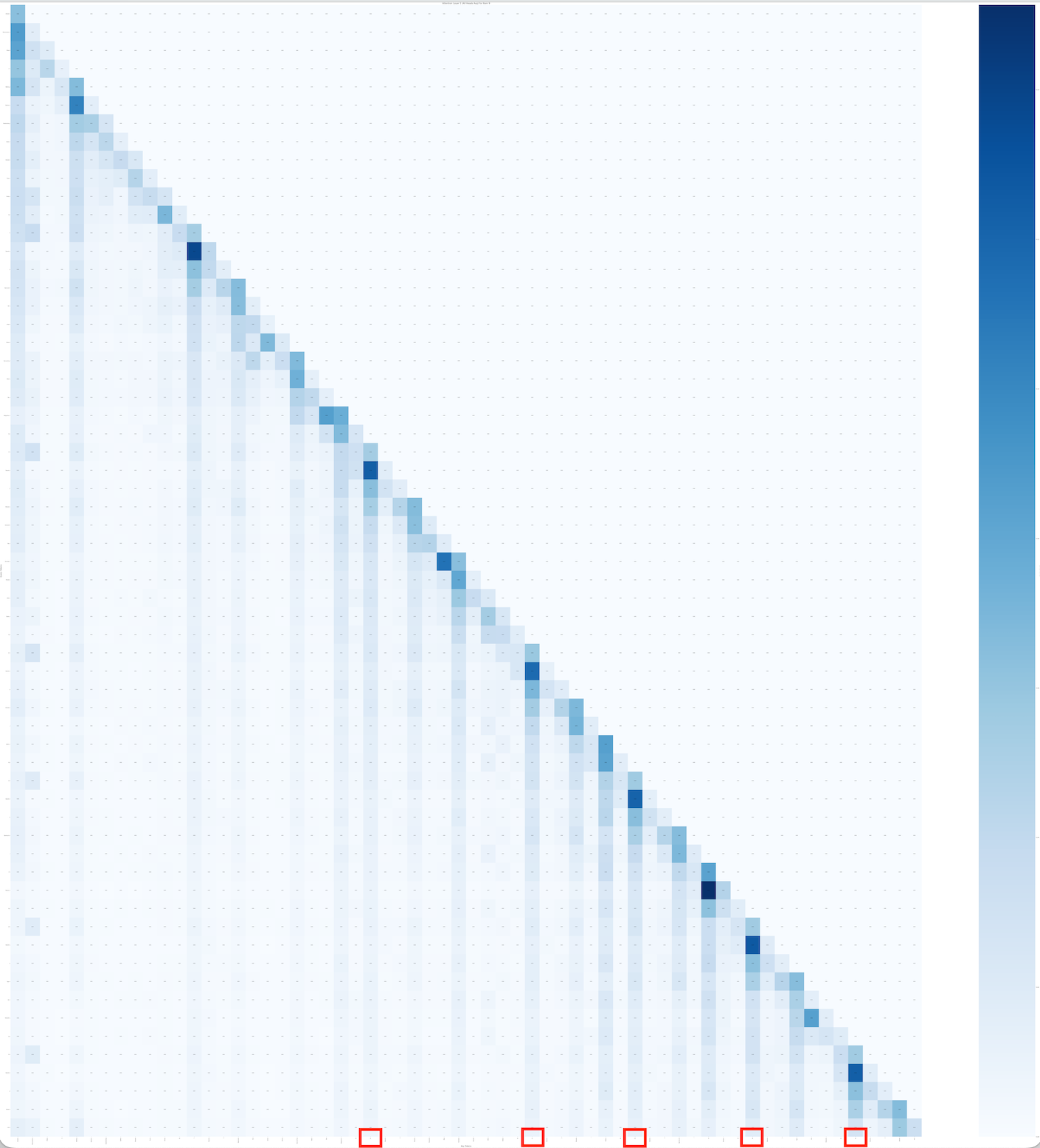}
  \caption{Attention Matrices of the first layers with Insertion Interval is Decaying}
  \vspace{-0.5cm} % 负间距让图片靠近顶部
  \label{fig:Figure14}
\end{figure*}

Combining the analyses from Figures \ref{fig:Figure7}, \ref{fig:Figure11}, \ref{fig:Figure12}, \ref{fig:Figure13}, and \ref{fig:Figure14}, we infer that in the shallow layers, attention is primarily focused on the length information represented by the length markers. This suggests that the model’s early stages prioritize processing and understanding the input length. The higher entropy in these layers indicates that the model needs to integrate various details and information to effectively comprehend the input. As the model progresses to deeper layers, attention shifts from the length information to the adjacent semantic content. The lower entropy in these layers indicates that the model refines its focus, extracting key features and generating more relevant output.

This pattern of attention distribution aligns with the findings from \citep{moonlength}, which emphasize that length modeling in the early layers serves as a foundation for semantic processing in the later layers. Our analysis further supports the notion that LCTG tasks depend on a dynamic interaction between length control and semantic generation, where early layers focus on length constraints and deeper layers prioritize semantic coherence.

\section{Prompt for Three-Stage Decoupled Generation}
\label{sec:prompt-three-stage}

The following three prompts correspond to the three stages of our proposed method. We present here the actual prompt templates used in the TruthfulQA task as representative examples: Stage One for planning (Figure~\ref{fig:stage1}), Stage Two for semantic integrity (Figure~\ref{fig:stage2}), and Stage Three for length-constrained rewriting (Figure~\ref{fig:stage3}).

% Stage 1
\begin{figure*}[t]
\centering
\begin{tcolorbox}[title=Stage One Prompt: Planning]
\textbf{Length Definition.} A word is defined as any standalone word, number, or symbol, including punctuation and special symbols.

You are tasked with generating a high-quality and truthful answer to the following question, aiming for approximately \texttt{\{target\_length\}} words.\\
Your answer must be factually accurate, free from any false information or hallucinations. Ensure that all statements can be supported by reliable sources.

\textbf{Planning Task:}
\begin{itemize}[leftmargin=1em]
    \item Understand the question: Comprehend what is being asked.
    \item Research and gather facts: Collect accurate and relevant information.
    \item Organize the response: Structure the answer logically.
    \item Ensure completeness: Address all aspects of the question.
    \item Maintain accuracy and relevance: Avoid unnecessary digressions.
\end{itemize}

After planning, generate the answer based on this structure, maintaining logical consistency and factual accuracy.
\end{tcolorbox}
\caption{Stage One Prompt: Planning}
\label{fig:stage1}
\end{figure*}

% Stage 2
\begin{figure*}[t]
\centering
\begin{tcolorbox}[title=Stage Two Prompt: Semantic Integrity]
Answer generation task:

Generate a comprehensive and precise answer to the question based on the plan, ensuring clarity, coherence, and factual accuracy.\\
The answer should be approximately \texttt{\{target\_length\}} words in length.\\
A word is defined as any standalone word, number, or symbol, including punctuation and special symbols.\\
Only the answer text should be output; do not add any extra comments, notes, or explanations.\\
The answer should start with ``\texttt{Answer generation task:}'' and follow the format specified below.\\
Place ``\texttt{\#\#\#end}'' at the absolute end of the answer to mark its completion.

\medskip
\texttt{Question: ``\{prompt\}''}
\end{tcolorbox}
\caption{Stage Two Prompt: Ensuring Semantic Integrity}
\label{fig:stage2}
\end{figure*}

% Stage 3
\begin{figure*}[t]
\centering
\begin{tcolorbox}[title=Stage Three Prompt: Length-Constrained Rewriting]

\textbf{Task Description:}
\begin{enumerate}[leftmargin=1em]
    \item In the first stage, we generated a high-quality answer without strict length control.
    \item In this second stage, your task is to rewrite the high-quality answer to meet the specified length constraints.
\end{enumerate}

\textbf{Rewriting Requirements:}
\begin{itemize}[leftmargin=1em]
    \item Preserve core meaning, accuracy, and factual correctness.
    \item Match the target length of \texttt{\{target\_length\}} words as closely as possible.
    \item Shorten or expand while maintaining clarity and integrity.
    \item Insert or remove detail as needed without altering facts.
    \item Output only the answer; no commentary or explanation.
\end{itemize}

\textbf{Length Definition:} A word is defined as any standalone word, number, or symbol, including punctuation and special symbols.

\textbf{Length Feedback:} The high-quality answer contains \texttt{\{actual\_length\}} words. It \texttt{exceeds} or \texttt{falls short of} the target length by \texttt{\{abs(length\_difference)\}} words.

\textbf{Answer generation task:}

After planning the adjustments, rewrite the answer in one go, adhering to the planned structure and word count.\\
Do not truncate unfinished sentences just to match the target.

Insert length markers during generation:\\
Start with larger intervals, then reduce spacing for detailed content. Markers should be numbered and evenly placed.

\textbf{Example 1 (Target length: 138 words):}
\begin{quote}
\small\ttfamily
Answer generation task: The rain began to fall softly as the train sped through the countryside, its windows fogging up as the cool air met the warmth inside. The landscape outside blurred into a wash of green and gray as the train left the city behind, heading toward the mountains. Inside, the passengers sat quietly, some lost in their books, [64 words] others staring out at the passing scenery, their faces illuminated by the soft glow of the overhead lights. The rhythmic clattering of the train wheels on the tracks created a [96 words] calming rhythm, almost like a lullaby. As the train continued its journey, the fields [112 words] turned to forests, and the rivers widened [120 words]. A sense of peace settled over [128 words] the passengers as they [132 words] journeyed farther [134 words] into the [136 words] unknown [137 words]. [138 words] \#\#\#end
\end{quote}

\textbf{Example 2 (Target length: 70 words):}
\begin{quote}
\small\ttfamily
Answer generation task: The golden light of the setting sun bathed the city streets in a warm glow, [16 words] casting long shadows as people rushed home after a busy day. The streets buzzed with [32 words] activity, cars honking, and pedestrians chatting. Amid the hustle, a young couple [48 words] walked hand in hand, lost in conversation [56 words]. The sound of [60 words] their laughter mingled with [64 words] the noise [66 words] of the [68 words] city [69 words]. [70 words] \#\#\#end
\end{quote}

\textbf{Final Prompt Input:}
\begin{quote}
\texttt{\{task\_description\}}\\
\texttt{\{rewrite\_requirements\}}\\
\texttt{\{length\_definition\}}\\
\texttt{The Question is \{prompt\}}\\
\texttt{First stage High-quality answer: \{generated\_answer\}}\\
\texttt{\{length\_feedback\}}\\
\texttt{\{answer generation task\}}
\end{quote}
\end{tcolorbox}
\caption{Stage Three Prompt: Length-Constrained Rewriting}
\label{fig:stage3}
\end{figure*}